\documentclass{article}

    \PassOptionsToPackage{numbers, sort, compress}{natbib}

\usepackage[preprint]{neurips_2025}

\usepackage[utf8]{inputenc} 
\usepackage[T1]{fontenc}    
\usepackage{hyperref}       
\usepackage{url}            
\usepackage{booktabs}       
\usepackage{amsfonts}       
\usepackage{nicefrac}       
\usepackage{microtype}      

\usepackage{xspace}
\usepackage{graphicx}

\usepackage{adjustbox}
\usepackage[table]{xcolor}         

\usepackage{pifont}

\usepackage{multirow}
\usepackage{amsmath}
\usepackage{float}

\newcommand{\methodname}{SIGHT-Fusion\xspace}
\newcommand{\taskname}{SIGHT\xspace}

\newcommand{\subfootnotesize}{\fontsize{7.5pt}{9pt}\selectfont}
\newcommand{\myparagraph}[1]{\noindent\textbf{#1.}}

\title{SIGHT: \underline{S}ynthesizing \underline{I}mage-Text Conditioned and \underline{G}eometry-Guided 3D \underline{H}and-Object \underline{T}rajectories}

\author{
Alexey Gavryushin \Envelope \\
ETHZ\\[-1ex]
\And
Alexandros Delitzas \\
ETHZ, MPI for Informatics\\[-1ex]
\And
Luc Van Gool \\
{ETHZ, INSAIT, KU Leuven}\\[-1ex]
\And
Marc Pollefeys \\
ETHZ, Microsoft\\[-2ex]
\And
Kaichun Mo \\
NVIDIA\\[-2ex]
\And
Xi Wang \\
ETHZ, TUM, INSAIT\\[-2ex]
}

\begin{document}

\maketitle
\vspace{-0.4cm}
\begin{figure}[ht]
  \centering
  \includegraphics[width=1.0\linewidth]{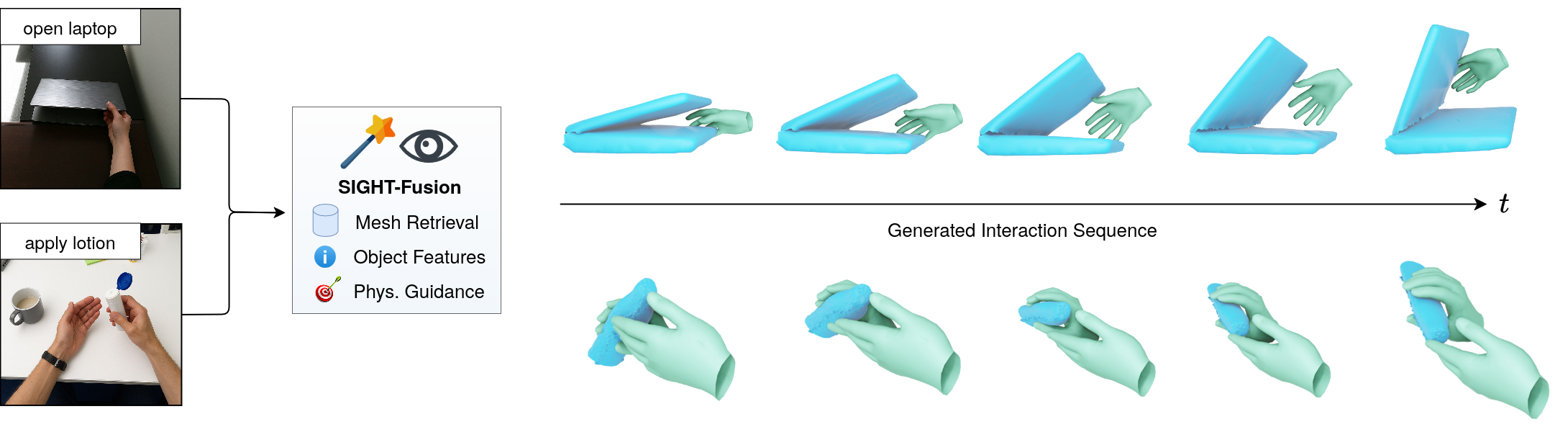} 
  \caption{\textbf{The proposed SIGHT task and SIGHT-Fusion method.}
  Given an input image showing an object being interacted with by a single hand or two hands as well as a task description, the SIGHT task is to generate realistic and physically plausible hand-object motion sequences mapping out possible trajectories of the hand(s) when interacting with the object as shown. 
  We propose a diffusion-based image-text conditioned method SIGHT-Fusion, which integrates a novel retrieval mechanism and inference-time guidance strategy to generate realistic and physically plausible 3D hand-object interaction trajectories.
  }
  \label{fig:teaser}
  \vspace{-3pt}
\end{figure}
\begin{abstract}
When humans grasp an object, they naturally form trajectories in their minds to manipulate it for specific tasks.
Modeling hand-object interaction priors holds significant potential to advance robotic and embodied AI systems in learning to operate effectively within the physical world.
We introduce SIGHT, a novel task focused on generating realistic and physically plausible 3D hand-object interaction trajectories from a single image and a brief language-based task description.
Prior work on hand-object trajectory generation typically relies on textual input that lacks explicit grounding to the target object, or assumes access to 3D object meshes, which are often considerably more difficult to obtain than 2D images.
We propose a novel diffusion-based image-text conditioned generative model that tackles this task by retrieving the most similar 3D object mesh from a database and enforcing geometric hand-object interaction constraints via a novel inference-time diffusion guidance.
We benchmark our model on the HOI4D and H2O datasets, adapting relevant baselines for this novel task. 
Experiments demonstrate our superior performance in the diversity and quality of generated trajectories, as well as in hand-object interaction geometry metrics.
We will publish our code and results. 
\end{abstract}
\section{Introduction}
As our hand grasps an object, we immediately plan out potential maneuvers to manipulate it for our intentions. 
%
%
Consider pouring some juice from a bottle into a cup -- it can be as straightforward as rotating your wrist. At a granular level, it requires a continuous adjustment of the hand translation and orientation to transfer exactly the desired amount of liquid into the target receptacle. 
Humans' hand motion planning systems are remarkably robust in adapting to unseen objects, emulating movements observed from others, and devising paths from visual cues and a downstream task objective in minds. 
Robotic agents and AI systems could greatly benefit from a similar capability, whether to anticipate human behavior, generate realistic animations, or facilitate interaction with the physical world.
%



In this paper, we propose a new task, SIGHT: given a single image showing a hand interacting with an object, along with a brief textual description of the task objective, the goal is to generate realistic and physically plausible 3D hand-object trajectories that meaningfully complete the action initiated in the image. 
This is a challenging task because the input is highly under-constrained—requiring the model to infer future motion from a single static frame and short text, while ensuring physical plausibility, contact consistency, and semantic alignment with the intended action.

Previous studies on hand-object interactions concentrate on detecting and segmenting pairs of hands and objects interacting within images~\cite{EKVisor, Zhang22EgoHOS}, or aim to reconstruct the 3D models of objects and the hands interacting with them from images~\cite{Hasson19LearningJointReconstruction, Ye22WhatsInYourHands} or videos~\cite{fan2024hold, ye2023vhoi}, but do not generate trajectories.
Furthermore, the field of human motion generation has hitherto focused on whole-body motion synthesis \cite{t2m, guo2020action2motion, MDM}, with little attention paid to synthesizing realistic, task-appropriate and interactive hand motions. 
%
In contrast, our focus is on generating dynamic 3D hand-object trajectories, which are crucial for manipulation in everyday human environments, yet using only a single static image and a short task description as inexpensive and easily accessible input.

We propose \methodname, a novel diffusion-based image-text conditioned generative model for tackling this challenging problem. 
Our framework first extracts textual features representing the task objective and visual features providing the grounded hand-object interaction scenario to condition the generative process. 
These features are fed into a motion diffusion network trained to produce continuous sequences of 3D hand and object poses, forming complete hand-object interaction trajectories. 
To encourage smoothness, we augment the standard diffusion objective with a velocity loss during training. 
At inference time, we introduce a novel geometry-based diffusion guidance mechanism to enforce physically plausible hand-object contact.
Specifically, we retrieve the most visually similar 3D object mesh to the input image from a database and incorporate a novel interpenetration loss into the test-time denoising process to guide geometry-aware trajectory generation.

We establish comprehensive baselines and evaluation metrics for the newly proposed \taskname task, leveraging the HOI4D~\cite{HOI4D} and H2O~\cite{H2O} datasets. 
Our evaluation spans a variety of settings, including different object categories, task types, single- or dual-hand interactions, and both rigid and articulated objects. 
We adapt several state-of-the-art motion generation methods to our setting as competitive baselines, and design a suite of quantitative metrics to assess trajectory accuracy, diversity, fidelity, as well as 3D hand-object interaction metrics such as interpenetration and contact consistency. 
Extensive experiments demonstrate that our approach generates more natural and diverse 3D interaction trajectories with realistic hand-object contacts, outperforming existing methods across both datasets. 
We further validate key design choices of our system, showing the effectiveness of different feature types, the proposed 3D retrieval augmentation, and the geometry-based diffusion guidance in enhancing trajectory quality.

In summary, the contributions of this paper are:
1) introducing a novel task of generating 3D hand-object interaction trajectories from a single image and a brief textural task description;
2) proposing a new diffusion-based conditional motion generative model that generates diverse and realistic 3D hand-object trajectories given the input text and image conditions;
3) introducing a novel geometry-based diffusion guidance at the inference time given a retrieved 3D object mesh to enforce physically plausible hand-object contact;
4) conducting experiments on two established datasets and demonstrating the superior performance of our proposed method compared to competitive baselines adapted to the new task.
We will release our code and model parameters to the public.
\vspace{1cm}
\section{Related Work}
\label{sec:related_work}
\vspace*{-0.2cm}
\myparagraph{Human motion generation}
Early work on human motion generation has largely focused on motion prediction~\cite{aksan2021spatio, caoHMP2020, Xiangbo2022recurrent, mix-and-match-perturbation} and unconstrained motion synthesis~\cite{ling2020character, wang2020adversarial, wang2020learning, yu2020structure, cai2021unified}. 
More recently, there has been growing interest in enhancing controllability in motion generation, leveraging various conditional signals such as text~\cite{karunratanakul2023gmd, kim2023flame, tevet2023human, MotionDiffuse, dabral2022mofusion}, audio~\cite{generatingHuman, dabral2022mofusion}, scene context~\cite{yi2024tesmo}, categorical actions~\cite{zhao2023modiff}, goal location~\cite{diomataris2024wandr} and motion of other people~\cite{liang2024intergen, shafir2024priormdm}.

Variational Autoencoders (VAEs) have been extensively used for generating realistic human motion from textual descriptions~\cite{petrovich2022temos, t2mgpt, generatingHuman, diomataris2024wandr}. 
Another line of work, such as MotionCLIP~\cite{tevet2022motionclip} and TM2T~\cite{tm2t}, employ transformer-based architectures~\cite{vaswani2017attention} to align the 3D human motion manifold with the semantically rich CLIP space~\cite{CLIP}, thus inheriting its capabilities.

Building on the success of diffusion models~\cite{ho2020ddpm, StableDiffusion, dhariwal2021diffusion} in generative modelling, 
several diffusion-based methods~\cite{MotionDiffuse, chen2023executing, tevet2023mdm, karunratanakul2023gmd, jiang2024motiongpt} have been proposed to tackle the task of text-based human motion generation. 
ReMoDiffuse~\cite{ReMoDiffuse} proposes a database retrieval mechanism to refine the denoising process. Specifically, the authors retrieve appropriate references from a database in terms of semantic and kinematic similarity and selectively leverage this information to guide the denoising process towards a more high-fidelity result.
This inspired the integration of a similarity-based retrieval module in our method. 
However, the aforementioned methods focus on synthesizing full-body motion, whereas hand-object interaction demands more fine-grained modeling of joint movements and precise alignment with object geometry.

\myparagraph{Hand-object interaction synthesis}
Similarly to 3D human motion synthesis, controllability has become a central objective in hand-object interaction synthesis, aiming to generate physically plausible and natural hand-object motion trajectories conditioned on various modalities.
Notably, IMoS~\cite{ghosh2022imos} synthesizes full-body motion and 3D object trajectories from text-based instruction labels using conditional variational auto-regressors to model the body part motions in an autoregressive manner.
ManipNet~\cite{manipnet} proposes a spatial hand-object representation that enables an autoregressive model to predict the hand poses given wrist and object trajectories as input.
Given an object geometry, an initial human hand pose as well as a sparse control sequence of object poses, CAMS~\cite{zheng2023cams} generates physically realistic hand-object manipulation sequences, using a c-VAE-based motion planner combined with an optimization module. 
Additionally, ManiDext~\cite{zhang2024manidext} adopts a hierarchical diffusion-based approach to synthesize hand-object manipulation based on 3D object trajectories. 
More recently, BimArt~\cite{zhang2025bimart} leverages distance-based contact maps to generate realistic bimanual motion, given an articulated 3D object along with its 6 DoF global states and 1 DoF articulation.
However, a key limitation of these methods is their reliance on 3D object motion sequences as input, which are often unavailable in real-world generative settings.
Another line of work~\cite{christen2022dgrasp, zhang2024graspxl, zhang2024artigrasp, jiang2021graspTTA} explores the generation of hand-object trajectories, yet requires running simulated environments to train policies using reinforcement learning, and primarily focuses on object grasping rather than modelling the full hand-object interaction.

In contrast to previous works, 
Text2HOI~\cite{Text2HOI} generates 3D motion for hand-object interaction, given a text and canonical object mesh as input. 
MACS~\cite{MACS2024} employs cascaded diffusion models to generate object and hand motion trajectories that plausibly adjust based on the object's mass and interaction type.
DiffH\textsubscript{2}O~\cite{christen2024diffh2o} further extends this direction by proposing a two-stage temporal diffusion process that decomposes hand-object interaction into distinct grasping and manipulation stages.
However, a key limitation of these methods is their reliance on 3D object shapes and their motion sequences as input, which are often unavailable in real-world generative settings.
In contrast, our method can generate realistic motion sequences from a single 2D image, without the need for 3D object data. 

\begin{figure}[t!]
    \centering
    \includegraphics[width=0.98\linewidth]{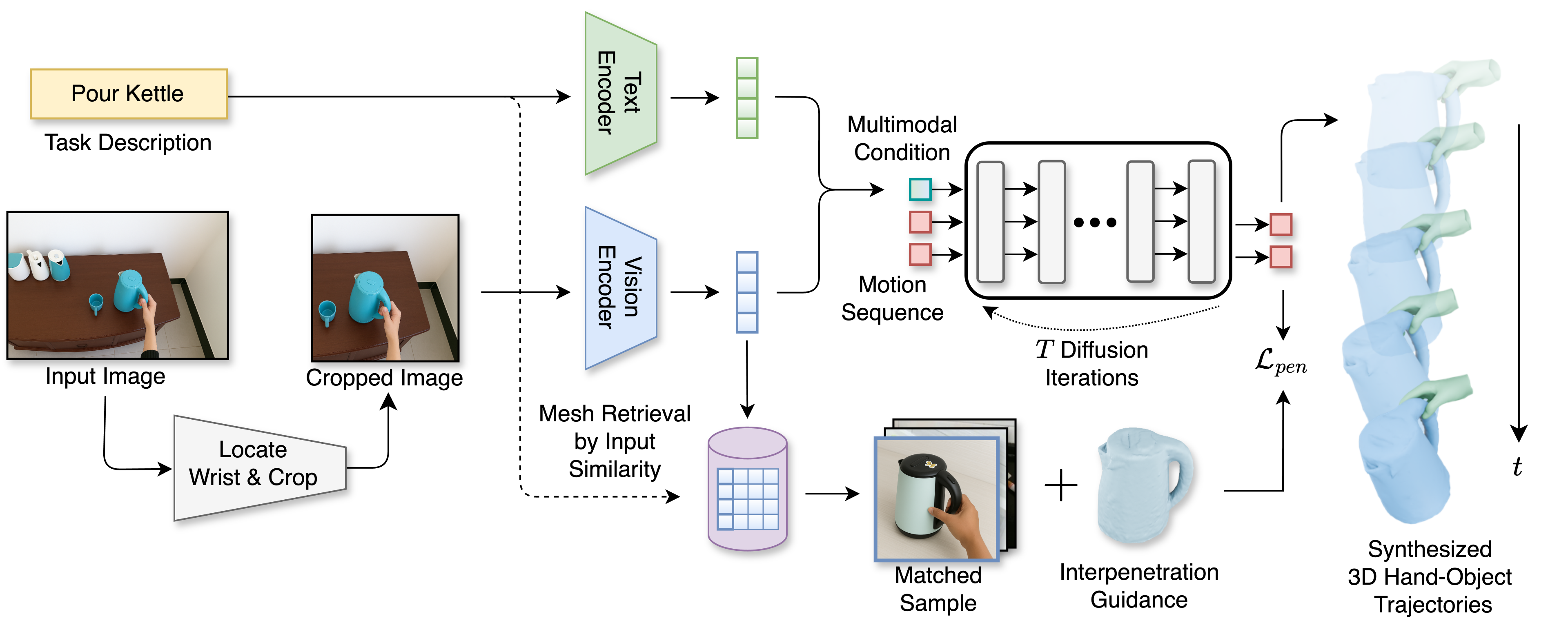}
    \caption{\textbf{An Overview of \methodname.} Given an input image, we first crop it to a region centered around the hand and the interacted object by detecting the wrist position. We then extract both textual and visual features from the input. These features are passed to a diffusion-based motion generator, which synthesizes realistic and task-appropriate 3D hand-object interaction trajectories. Additionally, we use the visual feature along with the task description to retrieve a corresponding 3D object, which serves as interpenetration guidance during inference.  
    }
    \label{fig:method}
\end{figure}
\vspace*{-0.2cm}
\section{Method}
\label{sec:method}
\vspace*{-0.2cm}

We start by formulating the \taskname task, and then describe the framework and loss formulation of the proposed \methodname method (see \autoref{fig:method} for an overview), designed to address \taskname by generating realistic, diverse, and physically plausible hand-object interaction trajectories from a single input image and a brief language-based task description. 

\subsection{Task Definition}
\label{sec:task}


We define the \taskname task as follows: The input to the hand-object interaction generator $\mathcal{M}$ is an image $\mathcal{I}$ showing a single hand or two hands enacting a certain action on an interacted object, such as \textit{lifting} or \textit{pouring} a kettle, and a brief textual description $\mathcal{T}$ of that action, consisting of a verb and a noun, e.g. \textit{pour kettle}. 
The goal is to generate a 3D motion sequence illustrating how the hand(s) and object move from the initial state depicted in the image. 
Specifically, the expected output is a sequence of hand poses and translations for one or both hands, denoted as $\mathcal{H}^s_{1:F} = \left(\left(\mathcal{H}^s_{P,1}, \mathcal{H}^s_{t,1}\right), \dots, \left(\mathcal{H}^s_{P,F}, \mathcal{H}^s_{t,F}\right)\right)$ as well as a sequence of object rotations and translations, $\mathcal{O}_{1:F} = \left(\left(\mathcal{O}_{R,1}, \mathcal{O}_{t,1}\right), \dots, \left(\mathcal{O}_{R,F}, \mathcal{O}_{t,F}\right)\right)$.  
$F$ is a predefined number of frames and $s\in\{left, right\}$ represents the hand side. 

We parameterize $\mathcal{H}^s_{P,i}$ using MANO \cite{MANO} and reasons about either a single hand in isolation, or both hands simultaneously for each frame, depending on the scenario. 
Each hand is parameterized by 16 joint rotations in $\mathcal{H}^s_{P,i}\in \mathbb{R}^{16\times D}$ and a 3D translation vector in $\mathcal{H}^s_{t,i}\in \mathbb{R}^3$ representing the respective wrist joint's offset from the coordinate origin. We normalize the translation such that the right hand's wrist is always at the origin in the first frame, i.e. $\mathcal{H}^{right}_{t,0} = \overrightarrow{0}_3$. $D$ is dependent on the rotation representation chosen. In our work, we encode all rotations using 6-dimensional representations of rotations, as suggested by \cite{ContinuityOfRotations}, yielding $D=6$.
Furthermore, we support reasoning about either rigid objects ($\mathcal{O}_{R,i} \in \mathbb{R}^{F \times D}$, $\mathcal{O}_{t,i} \in \mathbb{R}^{F \times 3}$) or objects with an additional articulated part ($\mathcal{O}_{R,i} \in \mathbb{R}^{F \times 2D}$, $\mathcal{O}_{t,i} \in \mathbb{R}^{F \times 6}$). This high-DoF articulation representation is preferred over explicit articulation parameters to accommodate rigid objects and enable a unified representation for both rigid and articulated objects. 
\subsection{Motion Network (\methodname)}
\label{sec:diffusion}

\myparagraph{Conditioning the trajectory generation}
\label{sec:conditioning}
We aim to provide the motion synthesis model with informative conditioning that helps to disambiguate the intended motion to be generated. While a textual description $\mathcal{T}$ of the task (e.g., \textit{pour kettle}) provides a coarse prior, we argue that the geometry of target object, as conveyed by its visual appearance, is still an important factor for synthesizing realistic trajectories with plausible contacts. Furthermore, we hypothesize that the pose and relative position of the hand in the input image $\mathcal{I}$ offer valuable clues about the starting configuration and initial steps of the motion. Hence, we choose to condition the model on both visual features and the textual description of the task.

Instead of simply providing the model with features from the whole image, we choose to extract features of the object being manipulated while minimizing irrelevant background information. 
We aim to extract these object features in a manner that is agnostic to the object category. Recognizing that the human hand is the common element across input images, we locate the wrist keypoint using an off-the-shelf model \cite{HaMeR}. We then crop a square region $\mathcal{I}_{crop}$ around the wrist to capture the hand and the object, without explicitly reasoning about the object itself. See \autoref{fig:method} for an example.  
%
We provide an analysis of the efficacy and failure cases of the proposed approach in the Supp.~Mat.

\myparagraph{Diffusion-based generation}
%
%
Given the shared requirement of reasoning about temporal sequences of both hand and object in 3D, we adopt state-of-the-art diffusion models for the \taskname task of generation of 3D hand-object interaction trajectories. 
%
%
We model the motion synthesis process over a sequence of $F$ frames, each having $J$ joints with $D$ pose features. For example, $J=2\times16+1$ represents two hands (16 joints each) and a rigid object, which we simply model as an additional joint. 
The synthesis process starts with a randomly sampled noise input $x_T \sim \left(\mathcal{N}(\vec{0}, I_{F\times J \times D}), \mathcal{N}(\vec{0}, I_{F\times J \times 3})\right)$ where $x_t \in \mathbb{R}^{F\times J \times D} \times \mathbb{R}^{F\times J \times 3}, \forall t \in \{1, ..., T\}$. 
Here, $T$ represents the total number of denoising steps to perform. 
Then a denoiser $\mathcal{G}$ is applied to iteratively denoise $x_T$ into the final motion $x_0$ in the $T$ diffusion steps.
%
%

For each input point $(\mathcal{I}_{crop}, \mathcal{T})$, we extract the text query feature $f_{\mathcal{T}} = E_{\mathcal{T}}({\mathcal{T}})$ and the visual feature $f_{\mathcal{I}_{crop}}=E_{\mathcal{I}_{crop}}(\mathcal{I}_{crop})$. 
Here, $ E_{\mathcal{T}}$ denotes the text encoder from the CLIP model~\cite{CLIP} and $E_{\mathcal{I}_{crop}}$ is a visual encoder based on SigLip~\cite{SigLIP}.
%
Let $c=(f_{\mathcal{T}}, f_{\mathcal{I}_{crop}})$ be the conditioning information 
and $\hat{x}_t$ represents the motion sequence generated in the diffusion step $t \in \{1, ..., T\}$ and $x_0$ the original sequence.
We initialize $\hat{x}_T$ by sampling random Gaussian noise. 
During training, we uniformly sample $t \in\{1, ... T\}$ to train $\mathcal{G}$, following DDPM \cite{ho2020ddpm}. 
During inference, for every step $t$, the denoising model $\mathcal{G}$ denoises $\hat{x}_{t+1}$ into $\hat{x}_{t}$ given conditioning $c$ and timestep $t$ as additional information. 
We adopt the Transformer decoder architecture proposed in \cite{ReMoDiffuse} for $\mathcal{G}$. 

To train our motion generator, we rely on a mean squared error (MSE) loss between the model's output and the ground-truth motion sequence. 
Specifically, we encode the object's rotation and translation $(\mathcal{O}_{R,i}, \mathcal{O}_{t,i})$ in each frame $i$ relative to the right wrist $(\mathcal{H}^{right}_{R,i,0}, \mathcal{O}_{t,i,0})$, i.e., the 0-th joint as defined by OpenPose~\cite{OpenPose}, within the same frame. We also employ a velocity loss to further reduce jitter in the generated sequences.
Our full loss formulation $\mathcal{L}$ consists of the DDPM-based simple reconstruction loss $\mathcal{L}_{rec}$ and a velocity loss $\mathcal{L}_{vel}$, which penalizes discrepancies in temporal differences for each joint:
\begin{align*}
& \mathcal{L} = \mathcal{L}_{rec} + \lambda_{vel}\mathcal{L}_{vel}, \\
& \mathcal{L}_{rec} = \mathbb{E}_{t~\sim\mathrm{Uniform}(\{1, ... T\})}  \left\lVert \hat{x}_t - x_0 \right\lVert_2^2, \\
& \mathcal{L}_{vel} = \mathbb{E}_{t~\sim\mathrm{Uniform}(\{1, ... T\})}  \sum_{f=1}^{F-1} \left\lVert \left(\hat{x}_{t,f} - \hat{x}_{t,f-1}\right) - \left(x_{0,f} - x_{0,f-1}\right)  \right\lVert_2^2
\end{align*}

where $\hat{x}_t = \mathcal{G}(\tilde{x}_t, t, c)$ and $\tilde{x}_t$ is an artificially noised version of the ground-truth motion $x_t$ used to train $\mathcal{G}$.
To avoid scenario-specific hyperparameter tuning, we set $\lambda_{vel} = 1$.

\myparagraph{Guiding the generation towards more realistic trajectories}
\label{sec:guidance}
The method discussed so far relies only on visual information and the prior from training to synthesize realistic trajectories. However, since it only works with a single monocular image, it has limited access to the object's geometry. 
Moreover, the absence of explicit constraints or loss terms that encourage realism
poses additional challenges in generating physically plausible motions. 
%

To aid the model in synthesizing more realistic trajectories, we use diffusion guidance~\cite{ClassifierGuidance} to steer the generation process toward more physically plausible outcomes. Diffusion Guidance provides an effective means of controlling the sample generated by a diffusion model by providing a gradient that is minimized at each diffusion step. This approach has been commonly used in text-conditioned image synthesis~\cite{ClassifierGuidance, epstein2023selfguidance}.
Specifically, at each diffusion step, we minimize a loss $\mathcal{L}_{pen}$ penalizing hand-object interpenetration:
\begin{equation*}
    \mathcal{L}_{pen} = \sum_{i=1}^F \, \sum_{v \in \mathcal{V}_i\cap\mathrm{Int}(M_i)} \min_{p \in \mathrm{\mathcal{V}_i}(M_i)} \| v - p \|_2^2,
\end{equation*}
where $\mathcal{M}_i$ is the mesh $\mathcal{M}$ of the interacted object at frame $i \in \{1, ..., F\}$ transformed according to the synthesized object motion $(\mathcal{O}_{R,i}, \mathcal{O}_{t,i})$. $\mathrm{Int}(M_i)$ denotes the space inside the mesh surface, and $\mathrm{\mathcal{V}_i}(M_i)$ are the vertices of the mesh. The set of MANO~\cite{MANO} hand vertices $\mathcal{V}_i = \mathrm{MANO}(\mathcal{H}_{P,i}, \mathcal{H}_{t,i})$ is from the synthesized hand poses $\mathcal{H}$ in the frame $i$.

Calculating $\mathcal{L}_{pen}$ requires access to the 3D object $\mathcal{M}$ during inference, which is not available as part of our input. 
We hence explore the strategy of using a database $\mathcal{D}$, where each entry consists of a key $k = (k_\mathcal{T}, k_{\mathcal{I}_{crop}})$ formed by a task descriptions $k_\mathcal{T}$ and visual features $k_{\mathcal{I}_{crop}}$ from the cropped region of interaction scenes, 
and a corresponding value $\mathcal{D}_k$, which contains the associated object mesh. 
The features $k_{\mathcal{I}_{crop}}$ are extracted from an image patch determined by the wrist position, following the same procedure used for the input image.

Given a new interaction defined by visual features $f_{\mathcal{I}_{crop}}$ and a task description $f_\mathcal{T}$, 
we query $\mathcal{D}$ to retrieve the object mesh whose visual features $k_{\mathcal{I}_{crop}}$ are most similar to $f_\mathcal{T}$, 
based on cosine similarity. 
%
Formally,
\begin{equation*}
    \mathcal{M} = \mathcal{D}_{k^*},\ k^* = {\arg\max}_{(k_{\mathcal{I}_{crop}}) \in \mathrm{keys}(\mathcal{D})} \frac{k_{\mathcal{I}_{crop}} \cdot f_{\mathcal{I}_{crop}}}{\lVert k_{\mathcal{I}_{crop}}\lVert \cdot \lVert f_{\mathcal{I}_{crop}} \lVert}
\end{equation*}
Additionally, we further filter the results by measuring the similarity between the retrieved task description $k_\mathcal{T}$ and the query task description $f_\mathcal{T}$. 
Note that this inference-time optimization approach is equally sensitive to meshes in the database as well as unseen ones, whereas a learning-based approach might generalize poorly to out-of-distribution objects. We provide experiments regarding the accuracy of our retrieval scheme and the alignment of the retrieved meshes with their visual observations shown depicted in the images in the Supp.~Mat. 

\section{Experiments}
\label{sec:exp}


In this section, we benchmark our model \methodname against base models on the proposed \taskname task. We evaluate the generated interaction trajectories based on their diversity, realism, and physical plausibility. 
Additionally, we find that visual features provide explicit grounding for interaction scenarios, and both retrieval and inference-time guidance lead to better motion generation.  

\subsection{Experimental Setup}
\label{sec:setup}

%
To establish a comprehensive evaluation for our newly introduced \taskname task, we adapt two hand-object interaction video datasets for our benchmark. 
We incorporate established metrics from the whole-body motion generation literature to evaluate the generated hand-object trajectories, and further assess the physical plausibility of the synthesized interactions using metrics measuring hand-object contact and interpenetration.

\myparagraph{Datasets}
The \textit{HOI4D} dataset~\cite{HOI4D} consists of first-person videos of hands interacting with everyday objects from 16 categories. 
3D scans of object instances are provided in the dataset, and 3D hand and 6D object poses corresponding to each video sequence are also available.  
The number of object instances varies within each category, ranging from 31 to 47, and the action tasks associated with these objects vary between 2 to 6 per category. 
Altogether, the dataset defines 31 action tasks. We merge and rename several tasks to extract 10 \textit{actions} from the original 31 tasks. Details of this grouping are provided in the Supp.~Mat. As there is no publicly available dataset split introduced for HOI4D, we define our own \textit{instance split} with the aim of testing cross-object-instance generalization ability for methods addressing the \taskname task.
%
We split each category action group into one subgroup with instances to use only for the training set, and another subgroup with instances to use only for the test set. 
Our proposed \textit{instance split} thus allows a meaningful evaluation of cross-instance action knowledge transfer within the same object category. It is also highly suitable for highlighting the benefits of our category-agnostic and mesh-independent method.
Simultaneously, more advanced reasoning is required for successful cross-instance transfer, as object instances in the test set may look different from those seen during training. 
%
A detailed description of the split is provided in the Supp.~Mat. 


The \textit{H2O} dataset~\cite{H2O} contains videos that capture two-hand interactions with eight different objects. It includes 36 distinct action classes along with rich annotations for 3D poses of the left and right hands and 6D object poses. 3D object meshes are also provided in the datset. We follow the official action split provided with the dataset.

\paragraph{Evaluation metrics.}
\label{sec:metrics}
%

For rigorous benchmarking, our \taskname task includes both hand trajectory-specific metrics, as well as metrics reasoning about the interaction of the hand and the object.
We measure the quality of the hand-object interaction trajectories generated by a method $M$ based on their \textit{accuracy} (ACC), \textit{diversity} (DIV), and \textit{fidelity to ground-truth} (FID). 
This combination of metrics is commonly used in the human motion generation literature \cite{guo2020action2motion, MDM}, as it measures the quality of motion sequences with respect to several desirable criteria. 

To measure the \textbf{accuracy} (ACC) of a motion generation method $M$, we consider the fraction of generated interaction trajectories a reference action classifier can correctly assign to the respective actions they were supposed to depict. The \textbf{diversity} (DIV) 
of $M$'s outputs is calculated through the Fréchet Inception distances (FIDs) \cite{FID} between action classifier features extracted from generated (gen.) and ground-truth (GT) motion groups. 
The diversity of the generated trajectories corresponds to the FID of one-half of a trajectory group to the other half. It is desirable in moderation: too little diversity corresponds to a method simply (re-)producing a few learned trajectories, while too much results in the hand trajectories becoming erratic. Hence, diversity scores closer to that of the GT trajectories are better. The \textbf{fidelity} (FID) of generated hand trajectories is calculated through the Fréchet Inception distances, similar to the diversity metric. 
Lower FIDs between generated and ground-truth sequences translate to a motion generation method producing trajectories more similar in distribution to real reference trajectories. 

We further investigate the quality of the generated hand-object interaction using \textit{interpenetration depth} (ID; in $cm$), interpenetration volume \textit{contact ratio} (IV; in $cm^3$) and \textit{contact ratio} (CR).
We report the largest \textbf{interpenetration depth} (ID), in centimeters, to which the hand(s) penetrated any of the objects. The maximum is taken over all frames of all sequences in the validation set. This metric is to be minimized to generate realistic motion sequences. The \textbf{interpenetration volume} (IV) measures the volume of the section of the hand penetrating the object, in cubic centrimeters and evaluated at the frame of the maximum interpenetration depth. This metric is a volumetric equivalent of the interpenetration depth. We additionally consider the \textbf{contact ratio} (CR) of hand vertices closer than some threshold $\tau$ to any vertex of the object, averaged over all frames and sequences. Here, we set $\tau = 5mm$. A contact ratio close to that of the ground truth is desirable, as the grasp is considered to mimic that in the ground-truth sequence more closely.

\paragraph{Implementation details.}
For each run, consistent with the human motion generation literature \cite{t2m, Action2Motion, MDM}, we select the checkpoint achieving the lowest FID metric on the test set, so as to produce the trajectories most similar to the test set trajectories. We average each score reported for HOI4D over 20 evaluations, and each score reported for the bimanual (and computationally more expensive) H2O over 5 evaluations for statistical robustness. Hyperparameters are provided in Supp.~Mat. 

\paragraph{Baselines.} 
We compare our method against three state-of-the-art baselines. As the proposed \taskname task is novel, we adapt existing baselines from the whole-body motion generation literature.

\emph{MDM}~\cite{MDM} trains a motion diffusion model using a transformer encoder backbone with classifier-free guidance. As a conditioning signal, it employs a CLIP-based textual embedding which is projected -- along with the diffusion timestep -- into the input token, together with the motion features to be denoised. Following the original formulation, we utilize the simple diffusion object augmented with the geometric losses (i.e., position and velocity) and retain all the standard hyperparameters. 

\emph{MotionDiffuse}~\cite{MotionDiffuse} is a diffusion-based model for motion generation that softly fuses CLIP encoded text features into the denoising process using cross-attention and stylization blocks. Following the original formulation, we use the simple diffusion objective and retain all the standard hyperparameters.

\emph{ReMoDiffuse}~\cite{ReMoDiffuse} is a retrieval-augmented motion diffusion model that conditions generation on semantically and kinematically similar motion trajectories. Given a textual description, it retrieves relevant motion samples from a database and feeds them into a semantics-modulated transformer, which extracts informative cues to guide the denoising process. We adopt the original text-conditioned loss formulation and use the standard hyperparameters.

\subsection{Evaluation of Generated 3D Hand-Object Interaction Trajectories}
\label{sec:evaluation}
\label{sec:results}

\newcommand{\hoiPerfGtACC}{1.000$^{\pm0.000}$}
\newcommand{\hoiPerfGtDIV}{11.782$^{\pm0.162}$}
\newcommand{\hoiPerfGtFID}{---}
\newcommand{\hoiPerfGtID}{2.742}
\newcommand{\hoiPerfGtIV}{14.981}
\newcommand{\hoiPerfGtCR}{0.063}

\newcommand{\hoiPerfMdmACC}{0.996$^{\pm0.002}$}
\newcommand{\hoiPerfMdmDIV}{\textbf{11.773}$^{\pm0.211}$}
\newcommand{\hoiPerfMdmFID}{\underline{0.132}$^{\pm0.016}$}
\newcommand{\hoiPerfMdmID}{3.471$^{\pm0.227}$}
\newcommand{\hoiPerfMdmIV}{35.145$^{\pm7.117}$}
\newcommand{\hoiPerfMdmCR}{0.074$^{\pm0.002}$}

\newcommand{\hoiPerfMotionDiffuseACC}{\underline{0.999}$^{\pm0.001}$}
\newcommand{\hoiPerfMotionDiffuseDIV}{11.613$^{\pm0.256}$}
\newcommand{\hoiPerfMotionDiffuseFID}{0.137$^{\pm0.018}$}
\newcommand{\hoiPerfMotionDiffuseID}{3.116$^{\pm0.187}$}
\newcommand{\hoiPerfMotionDiffuseIV}{31.033$^{\pm4.777}$}
\newcommand{\hoiPerfMotionDiffuseCR}{0.074$^{\pm0.003}$}

\newcommand{\hoiPerfReMoDiffuseACC}{0.991$^{\pm0.002}$}
\newcommand{\hoiPerfReMoDiffuseDIV}{11.702$^{\pm0.205}$}
\newcommand{\hoiPerfReMoDiffuseFID}{0.238$^{\pm0.031}$}
\newcommand{\hoiPerfReMoDiffuseID}{4.315$^{\pm0.158}$}
\newcommand{\hoiPerfReMoDiffuseIV}{72.595$^{\pm9.456}$}
\newcommand{\hoiPerfReMoDiffuseCR}{0.109$^{\pm0.003}$}

\newcommand{\hoiPerfTextToHOIACC}{0.952$^{\pm0.010}$}
\newcommand{\hoiPerfTextToHOIDIV}{10.751$^{\pm0.640}$}
\newcommand{\hoiPerfTextToHOIFID}{0.366$^{\pm0.091}$}
\newcommand{\hoiPerfTextToHOIID}{\underline{3.056}$^{\pm0.536}$}
\newcommand{\hoiPerfTextToHOIIV}{97.836$^{\pm58.466}$}
\newcommand{\hoiPerfTextToHOICR}{0.058$^{\pm0.001}$}

\newcommand{\hoiPerfOursACC}{\textbf{1.000}$^{\pm0.000}$}
\newcommand{\hoiPerfOursDIV}{\underline{11.820}$^{\pm0.204}$}
\newcommand{\hoiPerfOursFID}{\textbf{0.078}$^{\pm0.007}$}
\newcommand{\hoiPerfOursID}{\textbf{2.928}$^{\pm0.037}$}
\newcommand{\hoiPerfOursIV}{\textbf{23.108}$^{\pm7.339}$}
\newcommand{\hoiPerfOursCR}{\underline{0.069}$^{\pm0.002}$}

\newcommand{\hoiPerfImgMdmACC}{0.986$^{\pm0.002}$}
\newcommand{\hoiPerfImgMdmDIV}{11.653$^{\pm0.188}$}
\newcommand{\hoiPerfImgMdmFID}{0.153$^{\pm0.012}$}
\newcommand{\hoiPerfImgMdmID}{3.271$^{\pm2.360}$}
\newcommand{\hoiPerfImgMdmIV}{25.286$^{\pm3.828}$}
\newcommand{\hoiPerfImgMdmCR}{0.070$^{\pm0.003}$}

\newcommand{\hoiPerfImgModiffACC}{$1.000^{\pm0.000}$}
\newcommand{\hoiPerfImgModiffDIV}{$11.481^{\pm0.202}$}
\newcommand{\hoiPerfImgModiffFID}{$0.144^{\pm0.015}$}
\newcommand{\hoiPerfImgModiffID}{$2.995^{\pm0.836}$}
\newcommand{\hoiPerfImgModiffIV}{$28.269^{\pm9.353}$}
\newcommand{\hoiPerfImgModiffCR}{$\textbf{0.068}^{\pm0.002}$}

\newcommand{\hoiPerfImgReMoDiffuseACC}{1.000$^{\pm0.000}$}
\newcommand{\hoiPerfImgReMoDiffuseDIV}{11.744$^{\pm0.200}$}
\newcommand{\hoiPerfImgReMoDiffuseFID}{0.099$^{\pm0.008}$}
\newcommand{\hoiPerfImgReMoDiffuseID}{3.748$^{\pm0.136}$}
\newcommand{\hoiPerfImgReMoDiffuseIV}{74.467$^{\pm14.134}$}
\newcommand{\hoiPerfImgReMoDiffuseCR}{0.079$^{\pm0.003}$}

\newcommand{\htwoPerfGtACC}{0.856$^{\pm0.017}$}
\newcommand{\htwoPerfGtDIV}{12.397$^{\pm0.403}$}
\newcommand{\htwoPerfGtFID}{---}
\newcommand{\htwoPerfGtID}{3.452}
\newcommand{\htwoPerfGtIV}{13.562}
\newcommand{\htwoPerfGtCR}{0.084}

\newcommand{\htwoPerfMotionDiffuseACC}{\textbf{0.884}$^{\pm0.029}$}
\newcommand{\htwoPerfMotionDiffuseDIV}{12.689$^{\pm0.467}$}
\newcommand{\htwoPerfMotionDiffuseFID}{0.111$^{\pm0.022}$}
\newcommand{\htwoPerfMotionDiffuseID}{\textbf{3.542}$^{\pm0.231}$}
\newcommand{\htwoPerfMotionDiffuseIV}{36.041$^{\pm21.192}$}
\newcommand{\htwoPerfMotionDiffuseCR}{0.122$^{\pm0.002}$}

\newcommand{\htwoPerfMdmACC}{0.866$^{\pm0.021}$}
\newcommand{\htwoPerfMdmDIV}{\underline{12.595}$^{\pm0.330}$}
\newcommand{\htwoPerfMdmFID}{0.104$^{\pm0.005}$}
\newcommand{\htwoPerfMdmID}{3.729$^{\pm2.941}$}
\newcommand{\htwoPerfMdmIV}{\textbf{17.455}$^{\pm7.868}$}
\newcommand{\htwoPerfMdmCR}{\underline{0.112}$^{\pm0.007}$}

\newcommand{\htwoPerfReMoDiffuseACC}{\underline{0.879}$^{\pm0.024}$}
\newcommand{\htwoPerfReMoDiffuseDIV}{\textbf{12.514}$^{\pm0.117}$}
\newcommand{\htwoPerfReMoDiffuseFID}{\underline{0.100}$^{\pm0.002}$}
\newcommand{\htwoPerfReMoDiffuseID}{4.036$^{\pm0.075}$}
\newcommand{\htwoPerfReMoDiffuseIV}{21.035$^{\pm8.233}$}
\newcommand{\htwoPerfReMoDiffuseCR}{0.118$^{\pm0.010}$}

\newcommand{\htwoPerfTextToHOIACC}{0.421$^{\pm0.035}$}
\newcommand{\htwoPerfTextToHOIDIV}{12.436$^{\pm0.387}$}
\newcommand{\htwoPerfTextToHOIFID}{0.848$^{\pm0.194}$}
\newcommand{\htwoPerfTextToHOIID}{3.548$^{\pm0.091}$}
\newcommand{\htwoPerfTextToHOIIV}{95.212$^{\pm90.647}$}
\newcommand{\htwoPerfTextToHOICR}{0.102$^{\pm0.005}$}

\newcommand{\htwoPerfOursACC}{0.869$^{\pm0.008}$}
\newcommand{\htwoPerfOursDIV}{12.810$^{\pm0.566}$}
\newcommand{\htwoPerfOursFID}{\textbf{0.087}$^{\pm0.012}$}
\newcommand{\htwoPerfOursID}{\underline{3.618}$^{\pm0.113}$}
\newcommand{\htwoPerfOursIV}{\underline{20.111}$^{\pm5.172}$}
\newcommand{\htwoPerfOursCR}{\textbf{0.092}$^{\pm0.005}$}

{
\begin{table}
\caption{ \textbf{Quantitative results on the HOI4D dataset~\cite{HOI4D}.} Our method achieves the best performance compared to three state-of-the-art baselines. For comparison purposes, all methods use the same motion length of 196 frames. -I model variants use both the image and text modalities for conditioning. `$\rightarrow$' means results are better if the metric is closer to the real motions. We run all the evaluation 20 times for statistical robustness and `±' indicates the 95\% confidence interval. The best results are in \textbf{bold} and the second best results in \underline{underline}.} 
\centering
{
\subfootnotesize

\begin{tabular}{lcccccc}
\toprule

Method & ACC $\uparrow$ & DIV $\rightarrow$ & FID $\downarrow$ & ID (cm) $\downarrow$ & IV (cm$^3$) $\downarrow$ & CR $\rightarrow$ \\ \hline
\rowcolor[gray]{0.9} Ground-Truth & \hoiPerfGtACC & \hoiPerfGtDIV & \hoiPerfGtFID & \hoiPerfGtID & \hoiPerfGtIV & \hoiPerfGtCR \\
MDM \cite{MDM} & \hoiPerfMdmACC & \hoiPerfMdmDIV & \hoiPerfMdmFID & \hoiPerfMdmID & \hoiPerfMdmIV & \hoiPerfMdmCR \\
MotionDiffuse \cite{MotionDiffuse} & \hoiPerfMotionDiffuseACC & \hoiPerfMotionDiffuseDIV & \hoiPerfMotionDiffuseFID & \hoiPerfMotionDiffuseID & \hoiPerfMotionDiffuseIV & \hoiPerfMotionDiffuseCR \\
ReMoDiffuse \cite{ReMoDiffuse} & \hoiPerfReMoDiffuseACC & \hoiPerfReMoDiffuseDIV & \hoiPerfReMoDiffuseFID & \hoiPerfReMoDiffuseID & \hoiPerfReMoDiffuseIV & \hoiPerfReMoDiffuseCR \\ \hline 

MDM-I & \hoiPerfImgMdmACC & \hoiPerfImgMdmDIV & \hoiPerfImgMdmFID & \hoiPerfImgMdmID & \hoiPerfImgMdmIV & \hoiPerfImgMdmCR \\

MotionDiffuse-I & \hoiPerfImgModiffACC & \hoiPerfImgModiffDIV & \hoiPerfImgModiffFID & \hoiPerfImgModiffID & \hoiPerfImgModiffIV & \hoiPerfImgModiffCR \\

ReMoDiffuse-I & \hoiPerfImgReMoDiffuseACC & \hoiPerfImgReMoDiffuseDIV & \hoiPerfImgReMoDiffuseFID & \hoiPerfImgReMoDiffuseID & \hoiPerfImgReMoDiffuseIV & \hoiPerfImgReMoDiffuseCR \\ \hline


Ours & \hoiPerfOursACC & \hoiPerfOursDIV & \hoiPerfOursFID & \hoiPerfOursID & \hoiPerfOursIV & \hoiPerfOursCR \\
\bottomrule
\end{tabular}


}
\label{tab:baselines_hoi4d}
\end{table}
}

{
\begin{table}
\caption{ \textbf{Quantitative results on the H2O dataset~\cite{H2O}.} Our method achieves the best performance on the challenging FID and CR metrics, and comparable performance on the ID and IV metrics, compared with three state-of-the-art baselines. For comparison purposes, all methods use the same motion length of 196 frames. `$\rightarrow$' means results are better if the metric is closer to the real motions. We run all the evaluation 5 times for statistical robustness and `±' indicates the 95\% confidence interval. The best results are in \textbf{bold} and the second best results in \underline{underline}.} 
\centering
{
\subfootnotesize
\begin{tabular}{lcccccc}
\toprule

Method & ACC $\uparrow$ & DIV $\rightarrow$ & FID $\downarrow$ & ID (cm) $\downarrow$ & IV (cm$^3$) $\downarrow$ & CR $\rightarrow$ \\ \hline
\rowcolor[gray]{0.9} Ground-Truth & \htwoPerfGtACC & \htwoPerfGtDIV & \htwoPerfGtFID & \htwoPerfGtID & \htwoPerfGtIV & \htwoPerfGtCR \\
MDM \cite{MDM} & \htwoPerfMdmACC & \htwoPerfMdmDIV & \htwoPerfMdmFID & \htwoPerfMdmID & \htwoPerfMdmIV & \htwoPerfMdmCR \\
MotionDiffuse \cite{MotionDiffuse} & \htwoPerfMotionDiffuseACC & \htwoPerfMotionDiffuseDIV & \htwoPerfMotionDiffuseFID & \htwoPerfMotionDiffuseID & \htwoPerfMotionDiffuseIV & \htwoPerfMotionDiffuseCR \\
ReMoDiffuse \cite{ReMoDiffuse} & \htwoPerfReMoDiffuseACC & \htwoPerfReMoDiffuseDIV & \htwoPerfReMoDiffuseFID & \htwoPerfReMoDiffuseID & \htwoPerfReMoDiffuseIV & \htwoPerfReMoDiffuseCR \\ \hline 


 Ours & \htwoPerfOursACC & \htwoPerfOursDIV & \htwoPerfOursFID & \htwoPerfOursID & \htwoPerfOursIV & \htwoPerfOursCR \\
\bottomrule
\end{tabular}


\label{tab:baselines_h2o}
}
\end{table}
}

\newcommand{\condGtPerfACC}{1.000$^{\pm0.000}$}
\newcommand{\condGtPerfIV}{14.981}
\newcommand{\condGtPerfDIV}{11.782$^{\pm0.162}$}
\newcommand{\condGtPerfFID}{---}
\newcommand{\condGtPerfCR}{0.063}

\newcommand{\condTextPerfACC}{0.991$^{\pm0.002}$}
\newcommand{\condTextPerfIV}{72.595$^{\pm9.456}$}
\newcommand{\condTextPerfDIV}{11.702$^{\pm0.205}$}
\newcommand{\condTextPerfFID}{0.238$^{\pm0.031}$}
\newcommand{\condTextPerfCR}{0.109$^{\pm0.003}$}

\newcommand{\condImagePerfACC}{\underline{0.995}$^{\pm0.002}$}
\newcommand{\condImagePerfIV}{77.603$^{\pm13.352}$}
\newcommand{\condImagePerfDIV}{11.684$^{\pm0.189}$}
\newcommand{\condImagePerfFID}{0.106$^{\pm0.011}$}
\newcommand{\condImagePerfCR}{0.074$^{\pm0.002}$}

\newcommand{\condObjectPerfACC}{\textbf{1.000}$^{\pm0.000}$}
\newcommand{\condObjectPerfIV}{82.878$^{\pm10.054}$}
\newcommand{\condObjectPerfDIV}{\underline{11.831}$^{\pm0.198}$}
\newcommand{\condObjectPerfFID}{0.129$^{\pm0.012}$}
\newcommand{\condObjectPerfCR}{0.081$^{\pm0.002}$}

\newcommand{\condTextImagePerfACC}{\textbf{1.000}$^{\pm0.000}$}
\newcommand{\condTextImagePerfIV}{74.467$^{\pm14.134}$}
\newcommand{\condTextImagePerfDIV}{\textbf{11.744}$^{\pm0.200}$}
\newcommand{\condTextImagePerfFID}{0.099$^{\pm0.002}$}
\newcommand{\condTextImagePerfCR}{0.079$^{\pm0.003}$}

\newcommand{\condTextObjectPerfACC}{\textbf{1.000}$^{\pm0.000}$}
\newcommand{\condTextObjectPerfIV}{\underline{30.773}$^{\pm9.113}$}
\newcommand{\condTextObjectPerfDIV}{11.892$^{\pm0.197}$}
\newcommand{\condTextObjectPerfFID}{\underline{0.079}$^{\pm0.008}$}
\newcommand{\condTextObjectPerfCR}{\underline{0.071}$^{\pm0.005}$}

\newcommand{\condOursPerfACC}{\textbf{1.000}$^{\pm0.000}$}
\newcommand{\condOursPerfIV}{\textbf{23.108}$^{\pm7.339}$}
\newcommand{\condOursPerfDIV}{\textbf{11.820}$^{\pm0.204}$}
\newcommand{\condOursPerfFID}{\textbf{0.078}$^{\pm0.007}$}
\newcommand{\condOursPerfCR}{\textbf{0.069}$^{\pm0.002}$}

{
\centering
\begin{table}[t]
\caption{ \textbf{Ablation study on different conditioning signals.} Adding our proposed guidance term to the model, and combining text with object features yields the best performance on all metrics.} 
\centering
{
\subfootnotesize

\begin{tabular}{lcccccc}
\toprule

Method & ACC $\downarrow$ & DIV $\rightarrow$ & FID $\downarrow$ & IV (cm$^3$) $\downarrow$ & CR $\rightarrow$ \\ \hline
\rowcolor[gray]{0.9} Ground-Truth & \condGtPerfACC & \condGtPerfDIV & \condGtPerfFID & \condGtPerfIV & \condGtPerfCR \\
Text & \condTextPerfACC & \condTextPerfDIV & \condTextPerfFID & \condTextPerfIV & \condTextPerfCR \\
Scene Img.~ & \condImagePerfACC & \condImagePerfDIV & \condImagePerfFID & \condImagePerfIV & \condImagePerfCR \\
Object Img.~ & \condObjectPerfACC & \condObjectPerfDIV & \condObjectPerfFID & \condObjectPerfIV & \condObjectPerfCR \\
Text + Scene Img.~ & \condTextImagePerfACC & \condTextImagePerfDIV & \condTextImagePerfFID & \condTextImagePerfIV & \condTextImagePerfCR \\
Text + Object Img.~ & \condTextObjectPerfACC & \condTextObjectPerfDIV & \condTextObjectPerfFID & \condTextObjectPerfIV & \condTextObjectPerfCR \\
Text + Object Img. + Guidance~(Ours) & \condOursPerfACC & \condOursPerfDIV & \condOursPerfFID & \condOursPerfIV & \condOursPerfCR \\
\bottomrule
\end{tabular}


}
\label{tab:condition_ablation}
\end{table}
}

\newcommand{\gtPerfACC}{1.000$^{\pm0.000}$}
\newcommand{\gtPerfIV}{14.981}
\newcommand{\gtPerfDIV}{11.782$^{\pm0.162}$}
\newcommand{\gtPerfFID}{---}
\newcommand{\gtPerfCR}{0.063}

\newcommand{\gtMeshPerfACC}{1.000$^{\pm0.000}$}
\newcommand{\gtMeshPerfIV}{21.207$^{\pm6.234}$}
\newcommand{\gtMeshPerfDIV}{11.589$^{\pm0.152}$}
\newcommand{\gtMeshPerfFID}{0.076$^{\pm0.005}$}
\newcommand{\gtMeshPerfCR}{0.069$^{\pm0.004}$}

\newcommand{\noGuidancePerfACC}{\textbf{1.000}$^{\pm0.000}$}
\newcommand{\noGuidancePerfIV}{30.773$^{\pm9.113}$}
\newcommand{\noGuidancePerfDIV}{11.892$^{\pm0.197}$}
\newcommand{\noGuidancePerfFID}{\underline{0.079}$^{\pm0.008}$}
\newcommand{\noGuidancePerfCR}{0.071$^{\pm0.005}$}

\newcommand{\randomMeshPerfACC}{\textbf{1.000}$^{\pm0.000}$}
\newcommand{\randomMeshPerfIV}{\underline{24.132}$^{\pm7.347}$}
\newcommand{\randomMeshPerfDIV}{\textbf{11.802}$^{\pm0.227}$}
\newcommand{\randomMeshPerfFID}{\textbf{0.078}$^{\pm0.007}$}
\newcommand{\randomMeshPerfCR}{\underline{0.070}$^{\pm0.002}$}

\newcommand{\retrievedMeshPerfACC}{\textbf{1.000}$^{\pm0.000}$}
\newcommand{\retrievedMeshPerfIV}{\textbf{23.108}$^{\pm7.339}$}
\newcommand{\retrievedMeshPerfDIV}{\underline{11.820}$^{\pm0.204}$}
\newcommand{\retrievedMeshPerfFID}{\textbf{0.078}$^{\pm0.007}$}
\newcommand{\retrievedMeshPerfCR}{\textbf{0.069}$^{\pm0.002}$}

{
\centering
\begin{table}[t]
\caption{ \textbf{Ablation of guidance.} We study the effect of using our guidance term with different mesh types and validate its effectiveness in reducing interpenetration while letting the motions maintain a high diversity and fidelty to the ground-truth.} 
\centering
{
\small

\begin{tabular}{lcccccc}
\toprule

Method & ACC $\downarrow$ & DIV $\rightarrow$ & FID $\downarrow$ & IV (cm$^3$) $\downarrow$ & CR $\rightarrow$ \\ \hline
\rowcolor[gray]{0.9} Ground-Truth & \gtPerfACC & \gtPerfDIV & \gtPerfFID & \gtPerfIV & \gtPerfCR \\
\rowcolor[gray]{0.9} True Mesh & \gtMeshPerfACC & \gtMeshPerfDIV & \gtMeshPerfFID & \gtMeshPerfIV & \gtMeshPerfCR \\
No Guidance & \noGuidancePerfACC & \noGuidancePerfDIV & \noGuidancePerfFID & \noGuidancePerfIV & \noGuidancePerfCR \\
Rand.~Mesh in Category & \randomMeshPerfACC & \randomMeshPerfDIV & \randomMeshPerfFID & \randomMeshPerfIV & \randomMeshPerfCR \\
Retrieved Mesh (Ours) & \retrievedMeshPerfACC & \retrievedMeshPerfDIV & \retrievedMeshPerfFID & \retrievedMeshPerfIV & \retrievedMeshPerfCR \\
\bottomrule
\end{tabular}


}
\label{tab:guidance_ablation}
\end{table}
}


As evident in \autoref{tab:baselines_hoi4d}, our method outperforms all baselines in four of six metrics and performs favorably in the remaining two when evaluating on HOI4D. The object-image conditioning provides a strong signal about the object's geometry and its relationship to the hand, resulting in a significantly lower FID than all baselines. The advantages of our method further become evident on the ID, IV, and CD metrics, where our guidance term helps us achieve natural, low-penetration grasps.

As H2O includes only one object instance per task across 8 objects, models can rely on strong priors learned from the ground-truth sequences without needing to reason about object geometry. 
As shown in \autoref{tab:baselines_h2o}, although our method cannot unfold its full potential in such a setting, it still achieves the best average performance, outperforming all baselines strongly in FID and Contact Ratio while being the runner-up on Interpenetration Depth and Volume. We attribute this to the evaluation on H2O favoring methods that learn to reproduce training trajectories over those that reason about the object.

\begin{figure}[t!]
    \centering
    \includegraphics[width=1.0\linewidth]{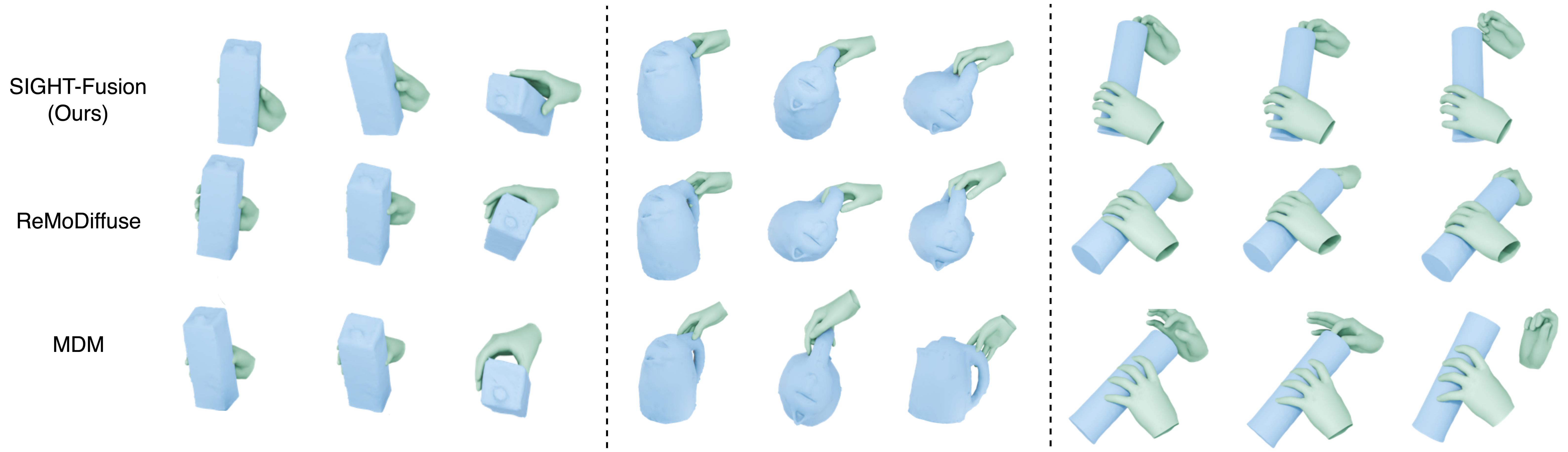}
    \vspace{-0.6cm}
    \caption{\textbf{Baseline Comparisons.} We provide qualitative comparisons of our method's synthesized motions (top row) to those of ReMoDiffuse (middle row) and MDM (bottom row), on objects from H2O (left, right) and HOI4D (center). The baselines exhibit interpenetration artifacts in the generated sequences, whereas our method produces realistic trajectories.
    }
    \label{fig:baseline_comparisons}
    \vspace{-0.3cm}
\end{figure}

Examples of single-handed (HOI4D) and bimanual (H2O) trajectories generated by our models, as well as comparisons to the ReMoDiffuse \cite{ReMoDiffuse} and MDM \cite{MDM} baselines, are visualized in \autoref{fig:baseline_comparisons}, confirming that our method achieves more physically realistic results while maintaining superior quantitative performance.

\paragraph{Ablations.}
\label{sec:ablations}
We present an ablation of several conditioning strategies for our model in \autoref{tab:condition_ablation}, starting with the text-based baseline \cite{ReMoDiffuse}. Generating motions based on text results in a lower accuracy while also markedly worsening the FID. Using whole-scene features by themselves improves these two metrics, with the FID reducing to less than half of the original value. Considering the object in isolation gives good accuracy and diversity, while slightly worsening the FID. Combining text with scene or object features gives us better FID and accuracy values than reasoning about them in isolation. Adding our guidance term to the model that combines text features with object features restores our method, giving the best performance on all metrics.

We further investigate different versions of our guidance term in \autoref{tab:guidance_ablation}, varying between using the ground-truth mesh, random meshes drawn from the training set and in the same category as the ground truth, as well as using meshes retrieved based on our proposed visual matching scheme. The results validate the need for the guidance term and further validate the usefulness of our feature-based matching over simply choosing a random mesh. Notably, while our guidance term modifies the generated motions to reduce physical artifacts, it does not worsen their trajectory-related metrics (ACC, DIV, FID), showing its usefulness for the investigated task.

\section{Discussion}
\label{sec:conclusion}

We introduce a novel task: the generation of natural and diverse 3D hand-object trajectories conditioned on single image inputs and language-based task descriptions.
To address this challenging problem, we propose a novel diffusion-based image-text-conditioned generative method \methodname.
%
To further improve the realism of generated trajectories, we integrate a visual similarity-based retrieval mechanism and a novel interpenetration guidance during inference. 
We set up comprehensive baselines adapted to the new task, and evaluation metrics on the HOI4D and H2O datasets. 
Experimental results demonstrate that our method outperforms baselines in terms of both diversity, naturalism, and physical plausibility. 
Ablation studies further validate the effectiveness of our proposed retrieval and inference-time guidance strategies and show how different input conditions influence the performance.

\myparagraph{Limitations and future directions}
While \methodname enables realistic and diverse 3D hand-object interaction motion generation, it relies on datasets that provide both videos of hand-object interaction sequences and the corresponding 3D object motions to construct the database.
Integrating existing image-to-3D generation pipelines could further improve the generalization ability of \methodname. 
Despite incorporating interpenetration guidance, the generated sequences may still exhibit hand-object penetrations. Exploring new formulations capable of generating realistic interaction sequences with valid hand-object contacts would further improve the method. 
We hope that our work will invite greater interest in this \taskname task, and that our generative framework will be of use in related applications, such as action anticipation and object manipulations in robotic settings.  

\myparagraph{Acknowledgements} 
This work was supported by the Swiss National Science Foundation Advanced Grant 216260: “Beyond Frozen Worlds: Capturing Functional 3D Digital Twins from the Real World”. AD is supported by the Max Planck ETH Center for Learning Systems (CLS).

\newpage

\maketitlesupp
\renewcommand{\thesection}{S\arabic{section}}
\renewcommand{\thesubsection}{S\arabic{section}.\arabic{subsection}}

\renewcommand{\thefigure}{S\arabic{figure}}

\renewcommand{\thetable}{S\arabic{table}}

\section{Mesh Retrieval}

\begin{table}[htbp]
\centering
\caption{Qualitative results of mesh retrieval}
\resizebox{\textwidth}{!}{%
\begin{tabular}{>{\centering\arraybackslash}m{0.16\textwidth}
                >{\centering\arraybackslash}m{0.16\textwidth}
                >{\centering\arraybackslash}m{0.16\textwidth}
                >{\centering\arraybackslash}m{0.16\textwidth}
                >{\centering\arraybackslash}m{0.16\textwidth}
                >{\centering\arraybackslash}m{0.16\textwidth}}

\toprule
Original image & Original mesh & Retrieved image & Retrieved mesh & Random image & Random mesh \\
\midrule

\includegraphics[width=\linewidth]{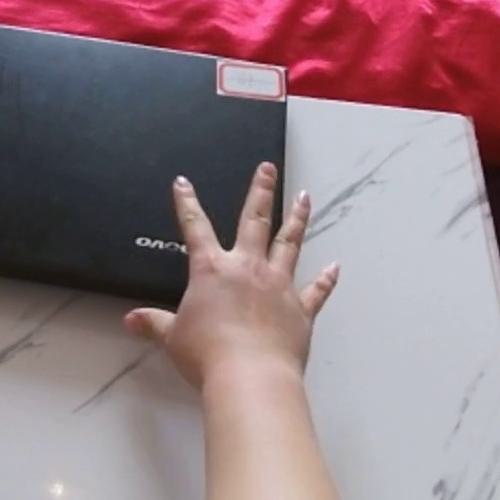} & 
\includegraphics[width=\linewidth]{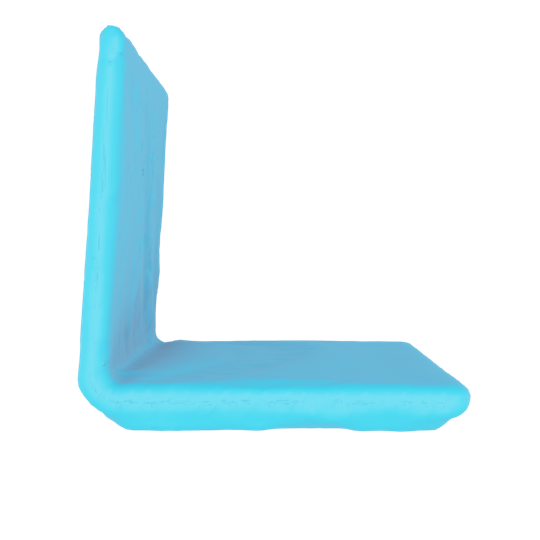} & 
\includegraphics[width=\linewidth]{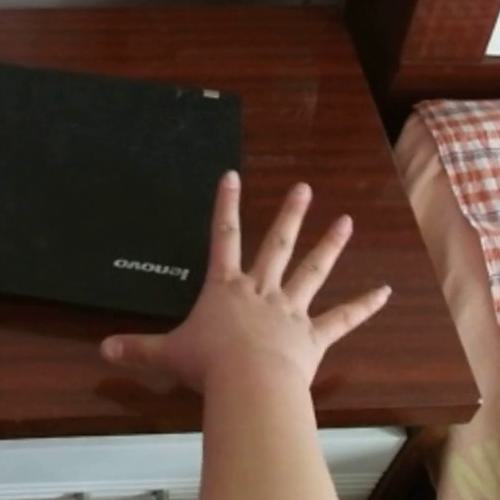} & 
\includegraphics[width=\linewidth]{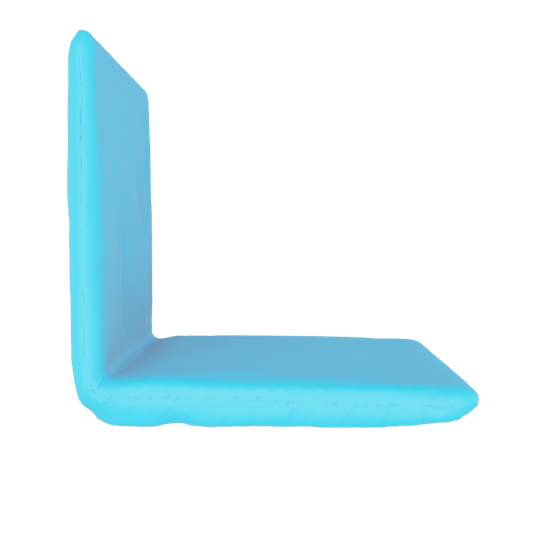} & 
\includegraphics[width=\linewidth]{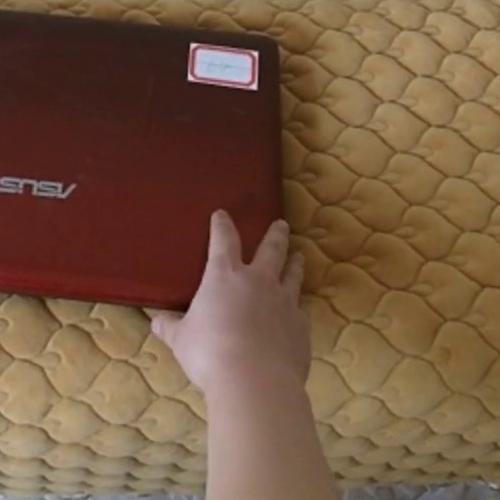} & 
\includegraphics[width=\linewidth]{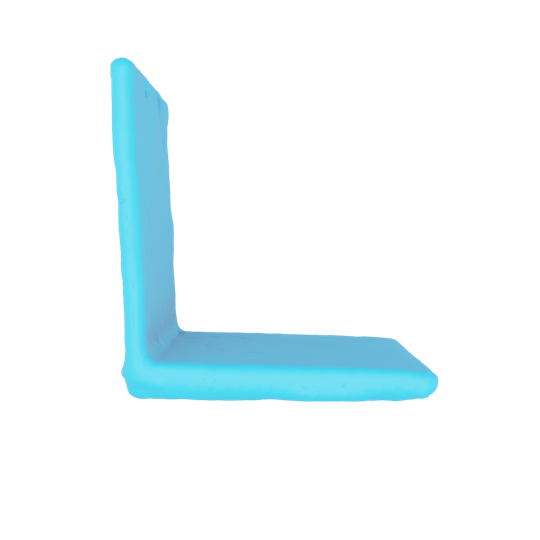} \\

\includegraphics[width=\linewidth]{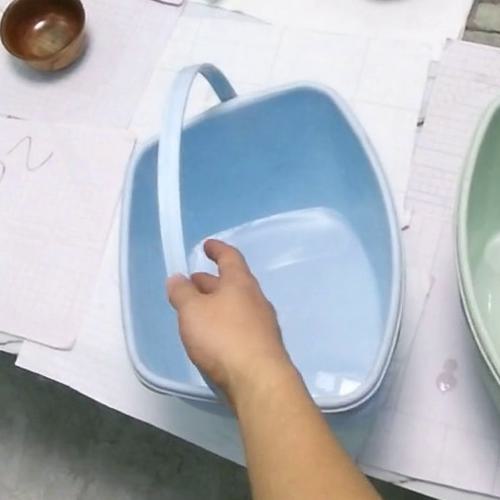} & 
\includegraphics[width=\linewidth]{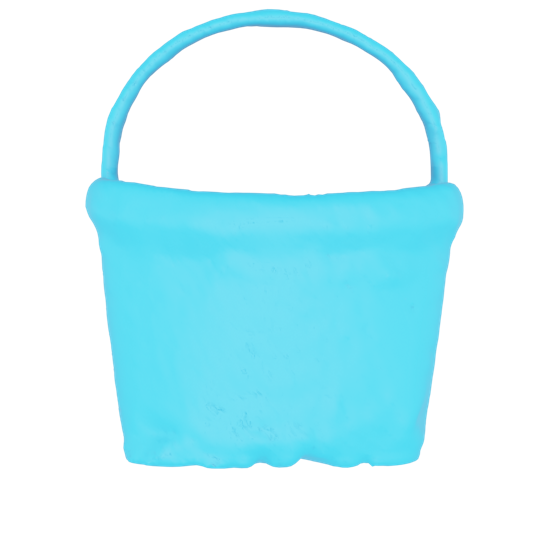} & 
\includegraphics[width=\linewidth]{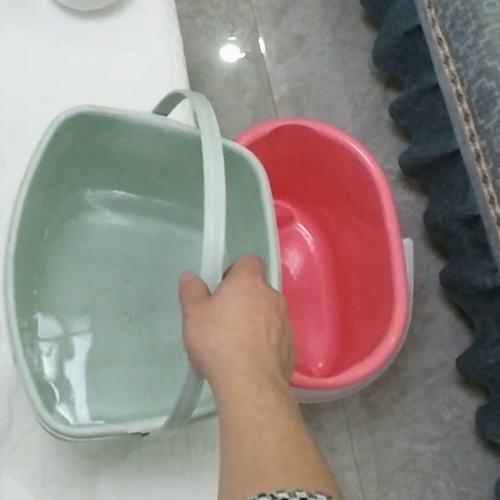} & 
\includegraphics[width=\linewidth]{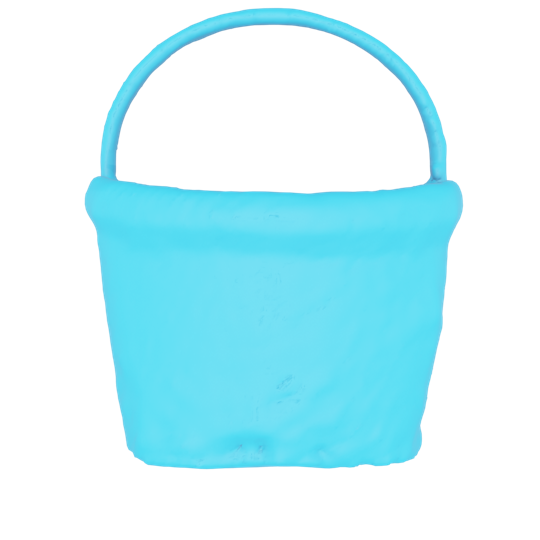} & 
\includegraphics[width=\linewidth]{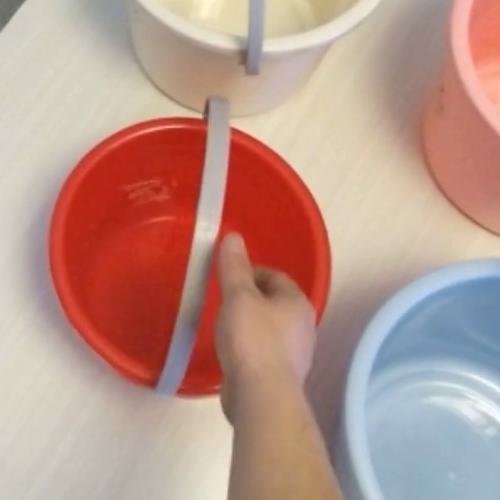} & 
\includegraphics[width=\linewidth]{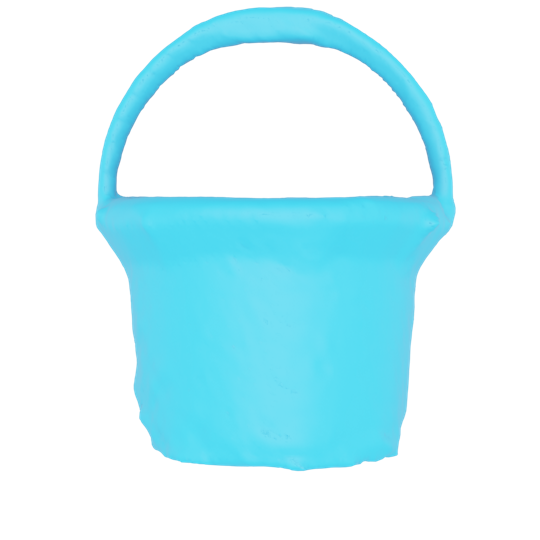} \\

\includegraphics[width=\linewidth]{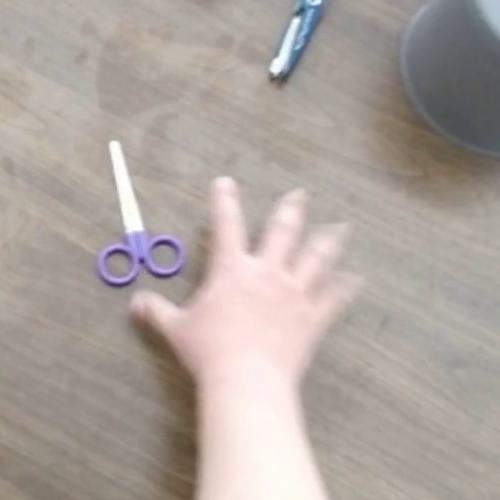} & 
\includegraphics[width=\linewidth]{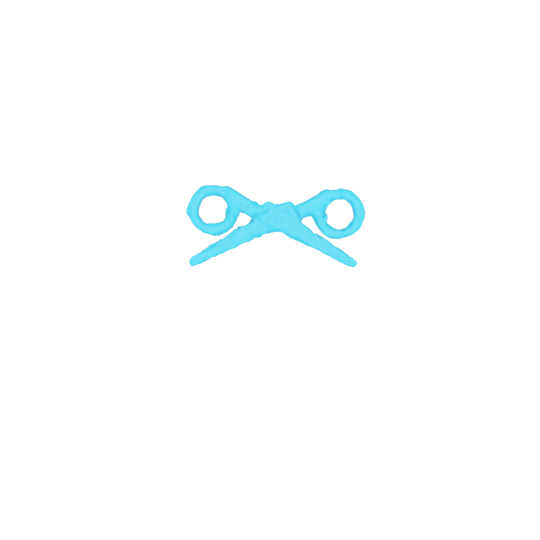} & 
\includegraphics[width=\linewidth]{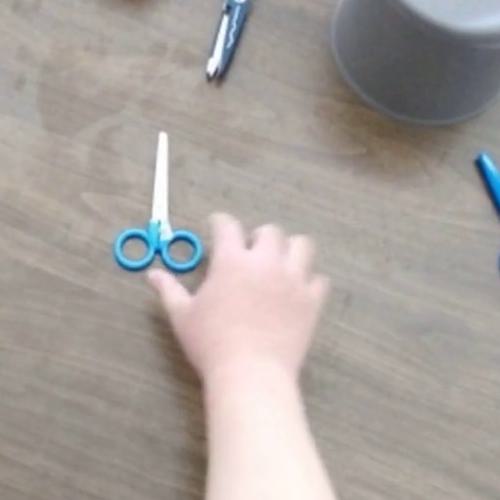} & 
\includegraphics[width=\linewidth]{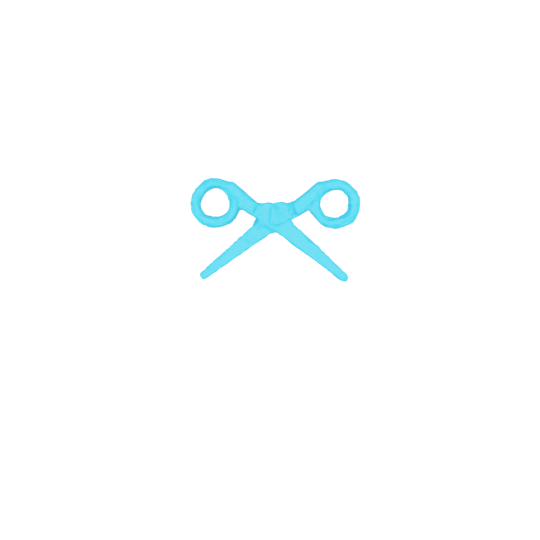} & 
\includegraphics[width=\linewidth]{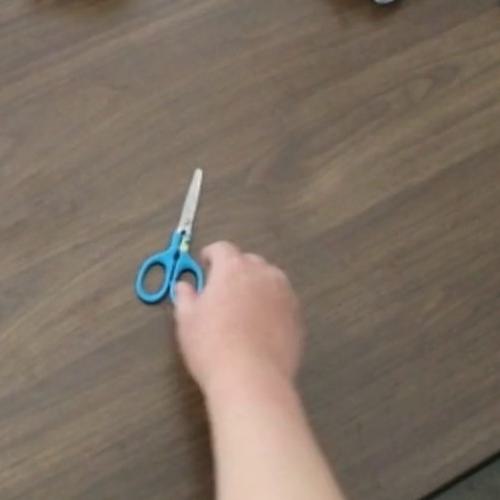} & 
\includegraphics[width=\linewidth]{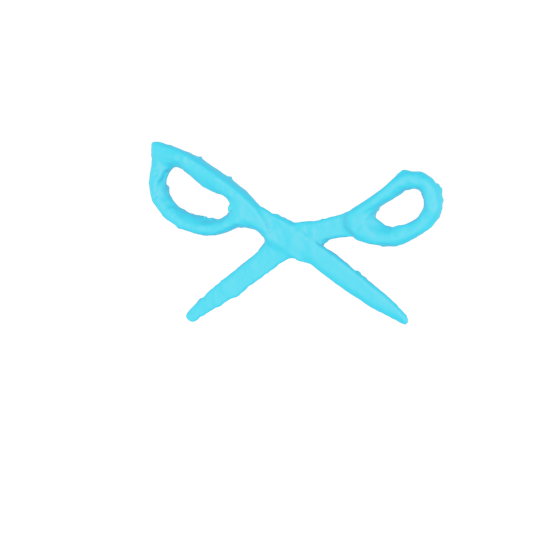} \\

\includegraphics[width=\linewidth]{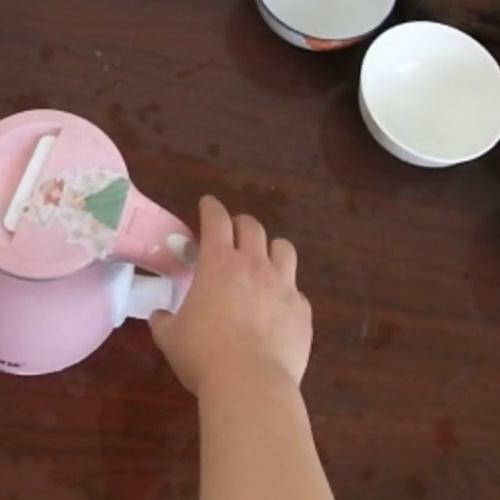} & 
\includegraphics[width=\linewidth]{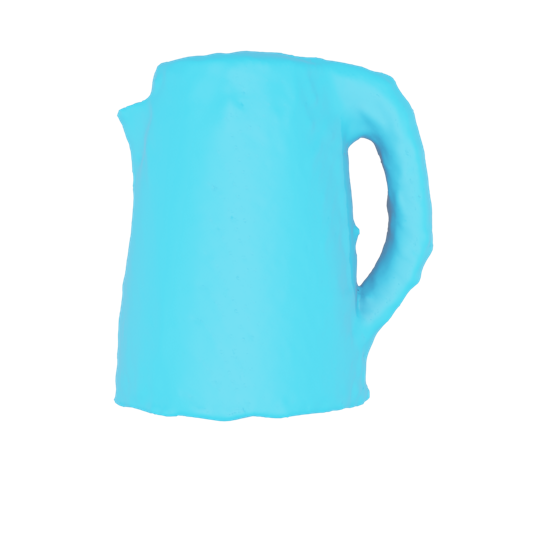} & 
\includegraphics[width=\linewidth]{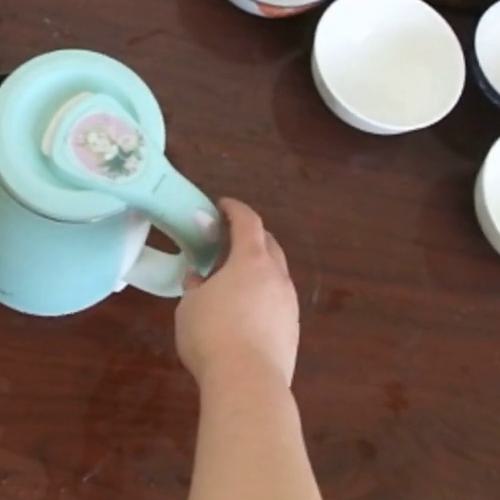} & 
\includegraphics[width=\linewidth]{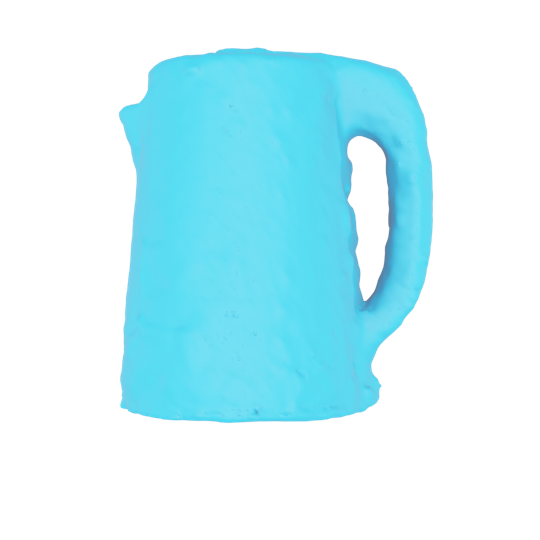} & 
\includegraphics[width=\linewidth]{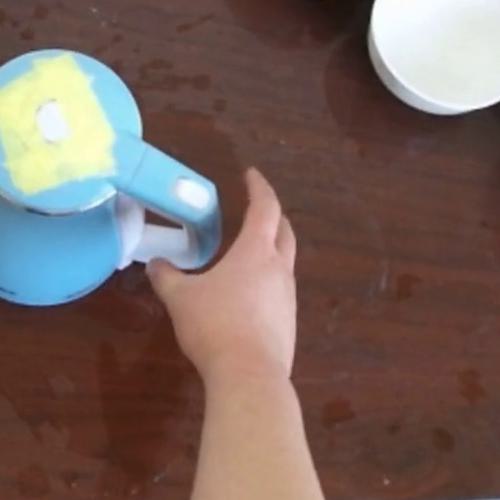} & 
\includegraphics[width=\linewidth]{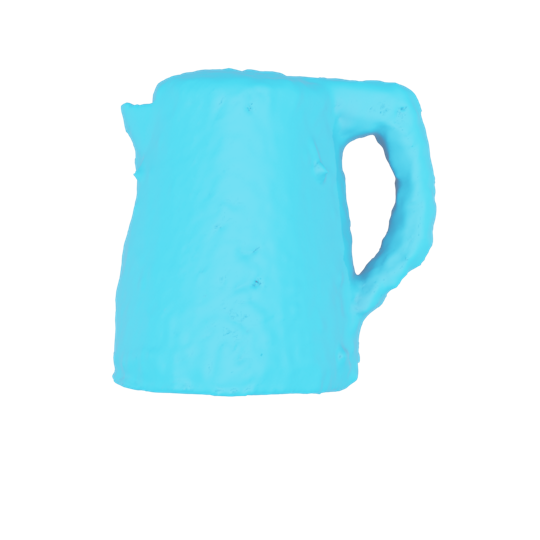} \\

\includegraphics[width=\linewidth]{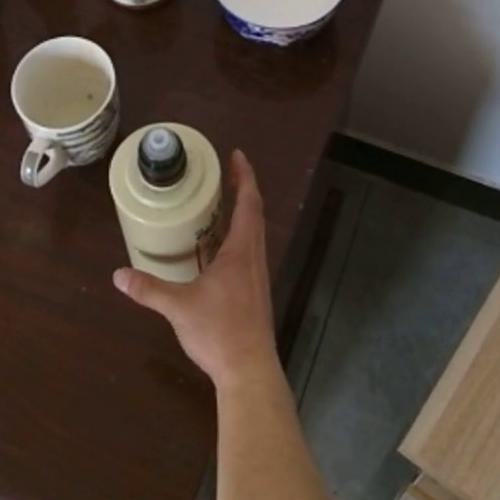} & 
\includegraphics[width=\linewidth]{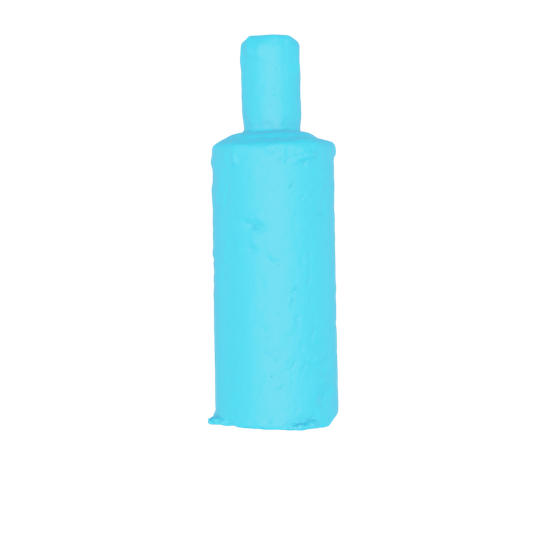} & 
\includegraphics[width=\linewidth]{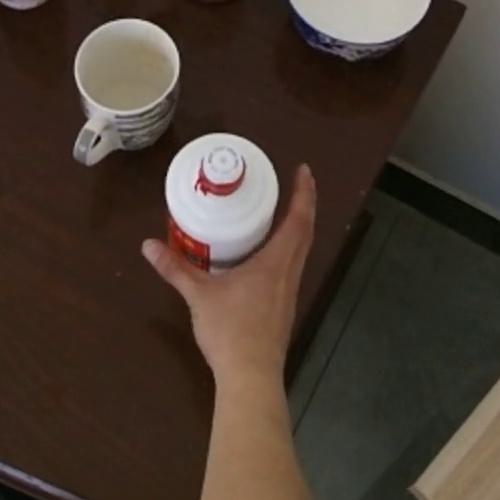} & 
\includegraphics[width=\linewidth]{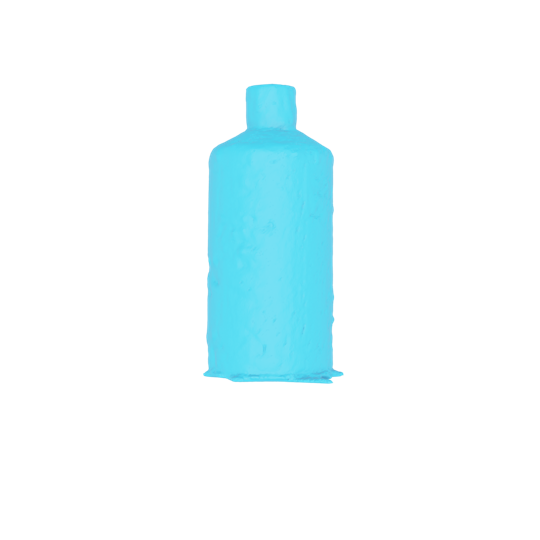} & 
\includegraphics[width=\linewidth]{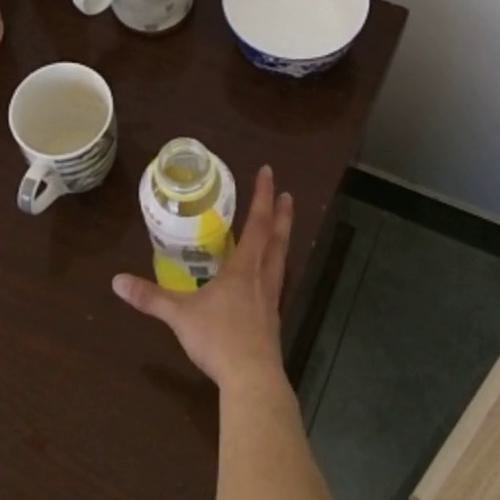} & 
\includegraphics[width=\linewidth]{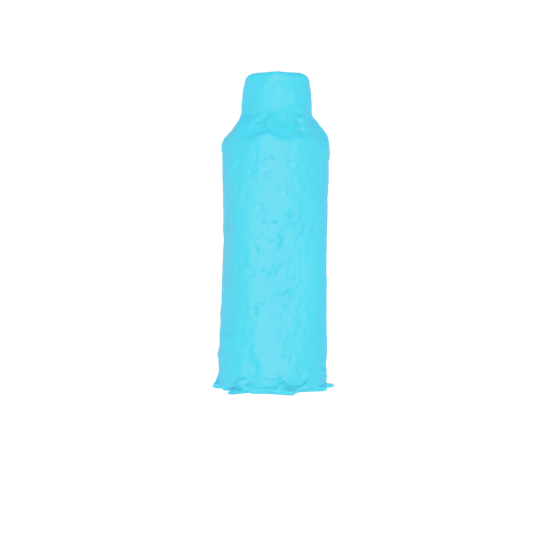} \\

\includegraphics[width=\linewidth]{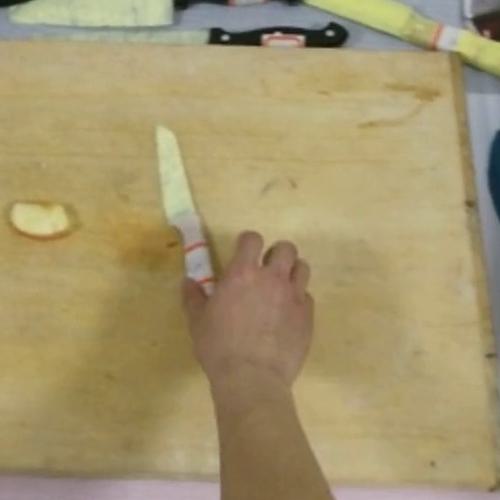} & 
\includegraphics[width=\linewidth]{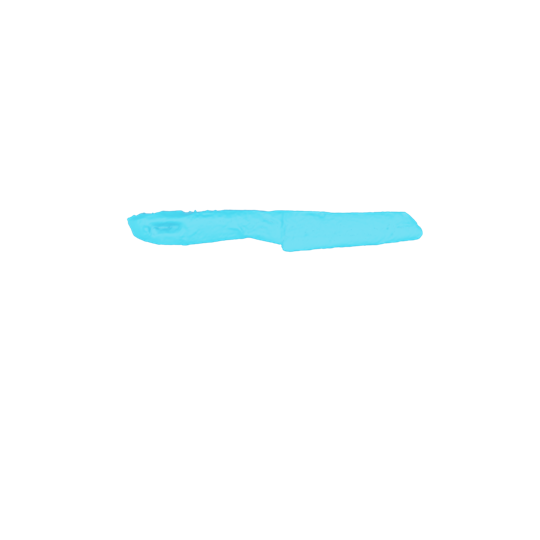} & 
\includegraphics[width=\linewidth]{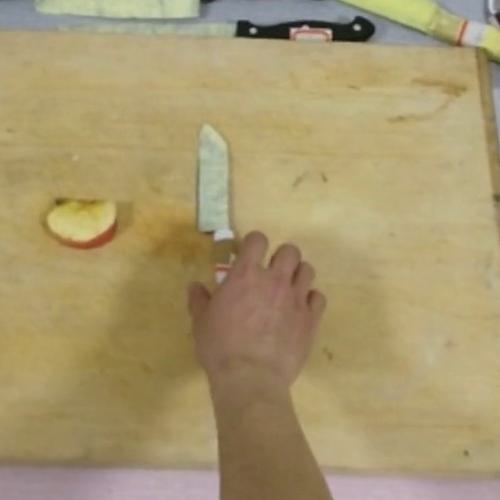} & 
\includegraphics[width=\linewidth]{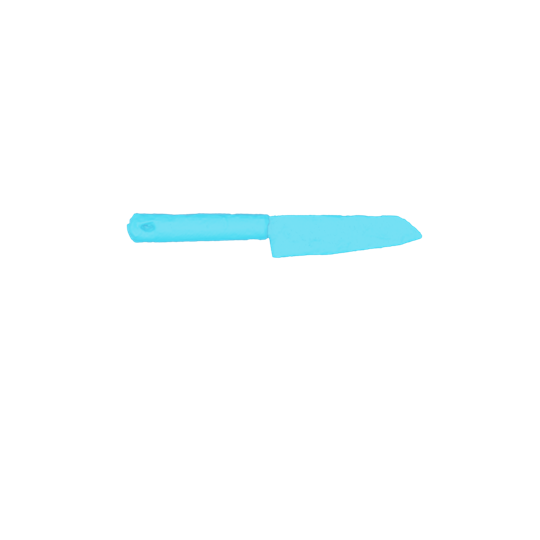} & 
\includegraphics[width=\linewidth]{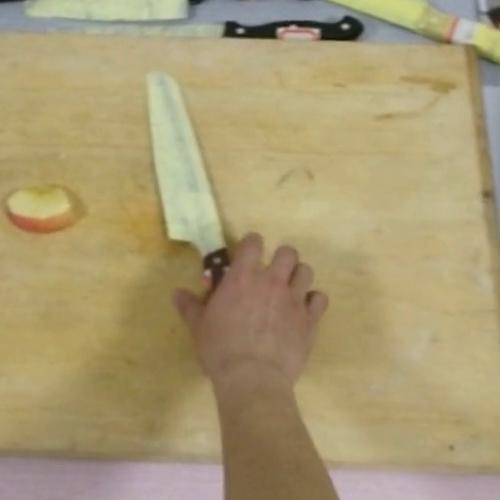} & 
\includegraphics[width=\linewidth]{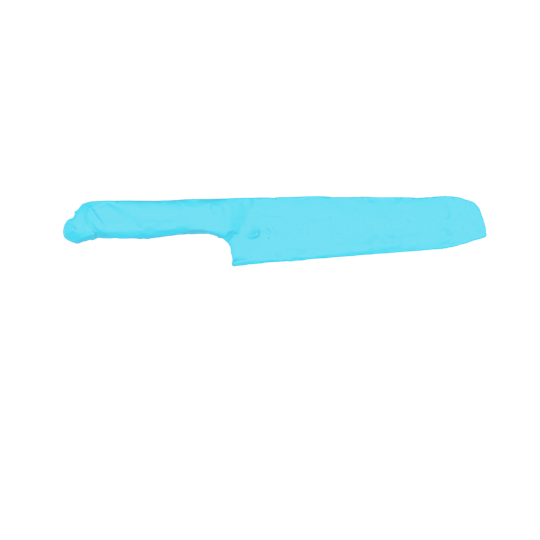} \\
\bottomrule
\end{tabular}
}
\label{fig:mesh_retrieval_vis}
\end{table}

Qualitative visualizations of validation set images/meshes, corresponding images/meshes retrieved from the training set for guidance-based optimization, as well as random images/meshes from the same category in the training set that were rejected in favor of the retrieved ones, are shown in \autoref{fig:mesh_retrieval_vis}.

\begin{figure}[H]
    \centering
   
  \renewcommand{\arraystretch}{0}
	
    \resizebox{\textwidth}{!}{
            \begin{tabular}{cccccc} 
\includegraphics[width=.90\linewidth]{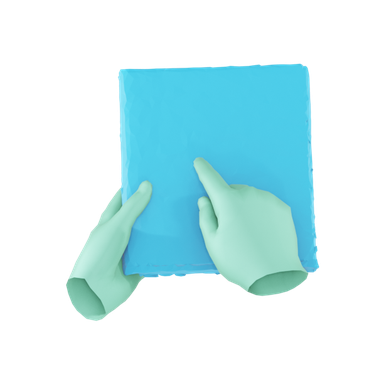}
\includegraphics[width=.90\linewidth]{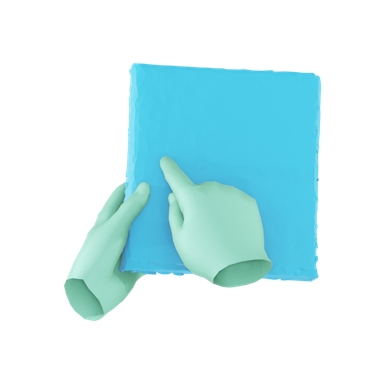}
\includegraphics[width=.90\linewidth]{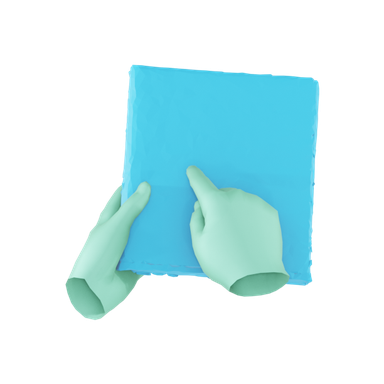}
\includegraphics[width=.90\linewidth]{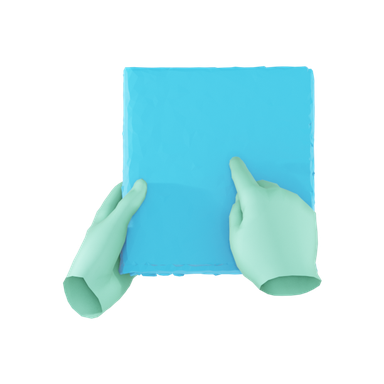}
\includegraphics[width=.90\linewidth]{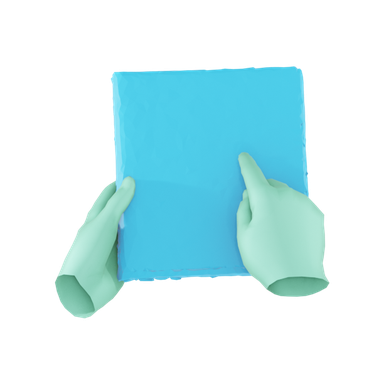}
\includegraphics[width=.90\linewidth]{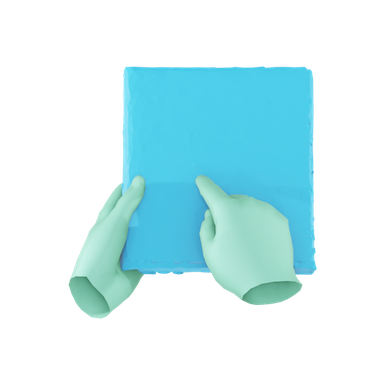} \\ \vspace{0mm} \vspace{-20mm} 
\includegraphics[width=.90\linewidth]{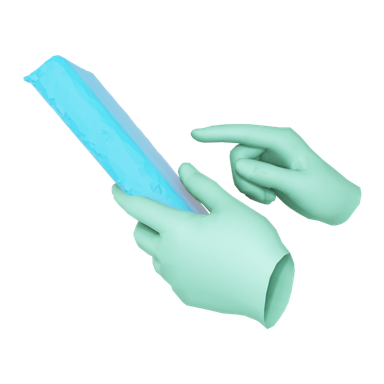}
\includegraphics[width=.90\linewidth]{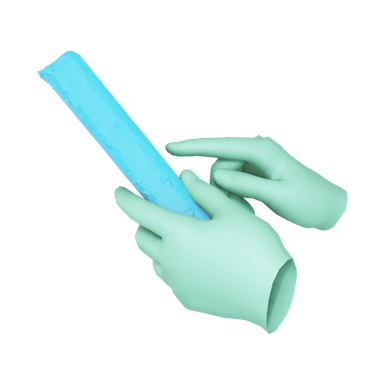}
\includegraphics[width=.90\linewidth]{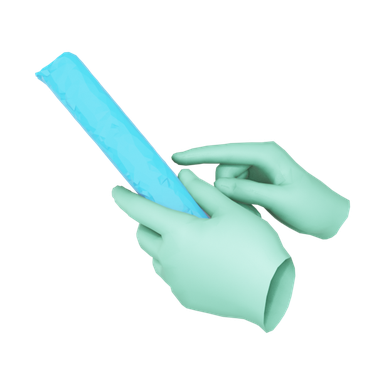}
\includegraphics[width=.90\linewidth]{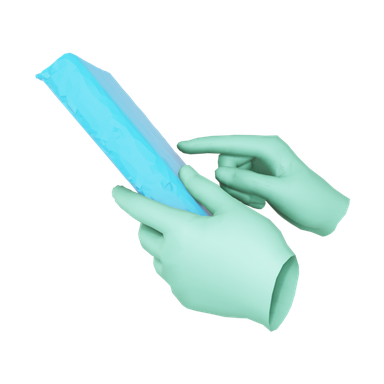}
\includegraphics[width=.90\linewidth]{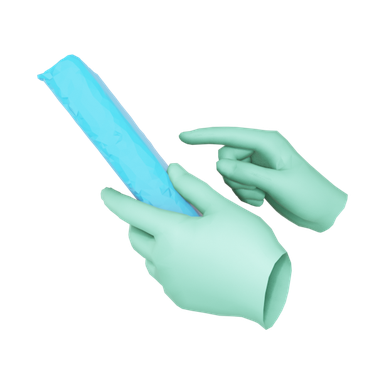}
\includegraphics[width=.90\linewidth]{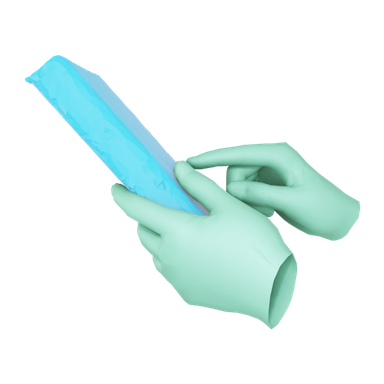}\\ \vspace{0mm} 
\includegraphics[width=.90\linewidth]{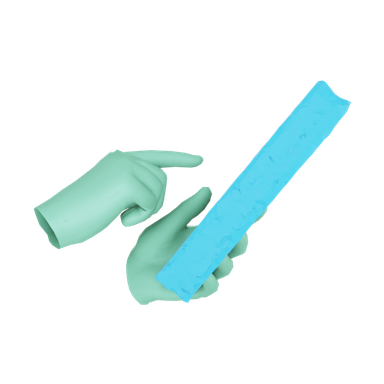}
\includegraphics[width=.90\linewidth]{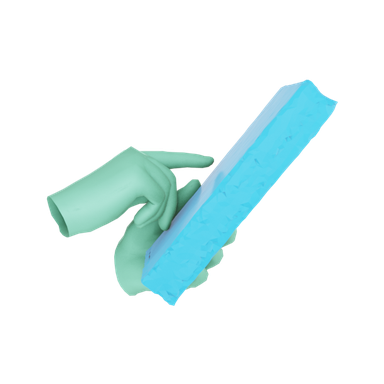}
\includegraphics[width=.90\linewidth]{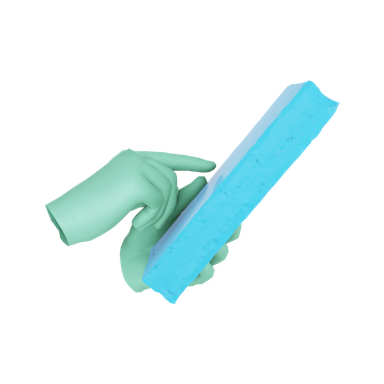}
\includegraphics[width=.90\linewidth]{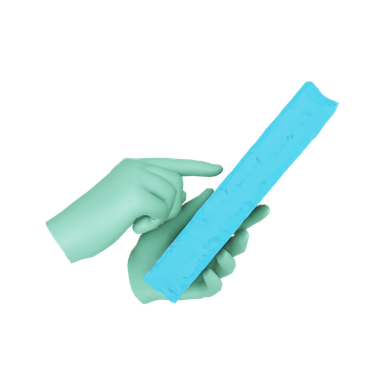}
\includegraphics[width=.90\linewidth]{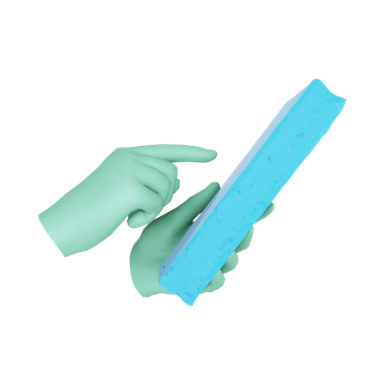}
\includegraphics[width=.90\linewidth]{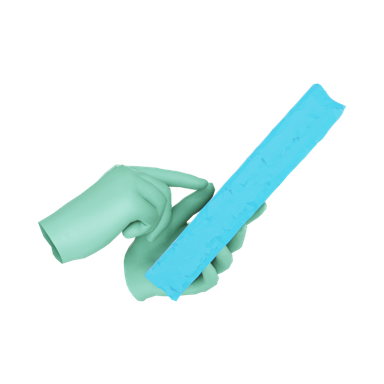} \\ \vspace{2mm} 

            \end{tabular}}
    
    \vspace{0.35cm}

    \resizebox{\textwidth}{!}{
            \begin{tabular}{cccccc} 
\includegraphics[width=.90\linewidth]{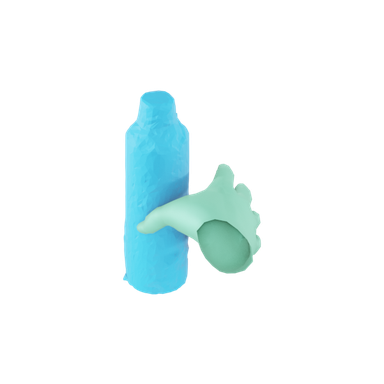}
\includegraphics[width=.90\linewidth]{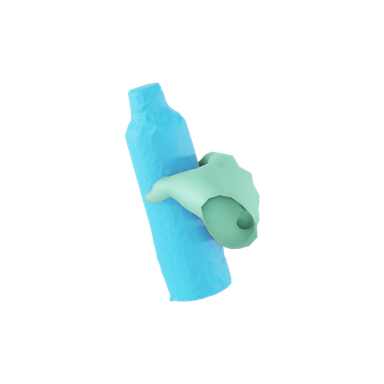}
\includegraphics[width=.90\linewidth]{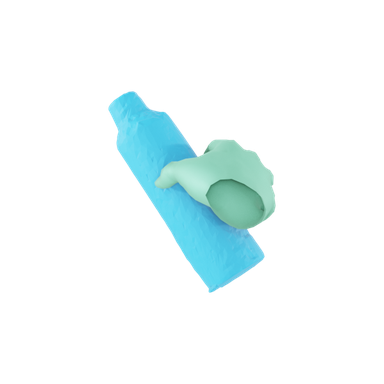}
\includegraphics[width=.90\linewidth]{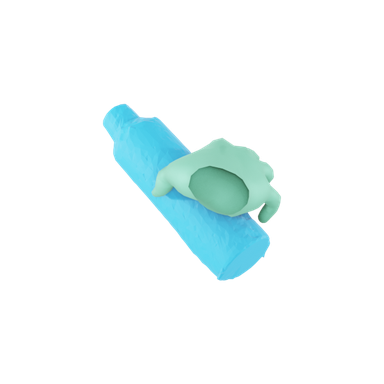}
\includegraphics[width=.90\linewidth]{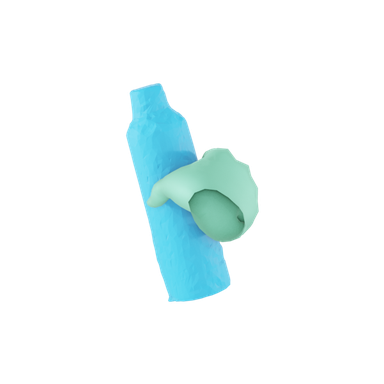}
\includegraphics[width=.90\linewidth]{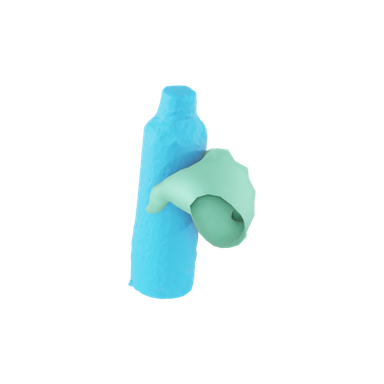} \\ \vspace{0mm} 
\includegraphics[width=.90\linewidth]{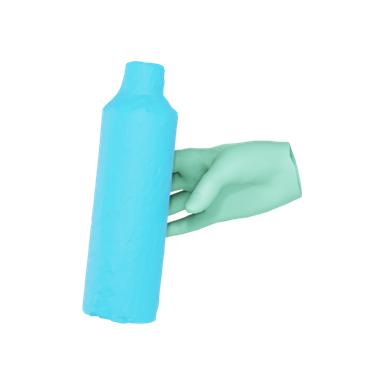}
\includegraphics[width=.90\linewidth]{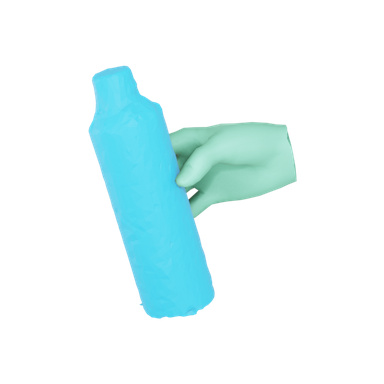}
\includegraphics[width=.90\linewidth]{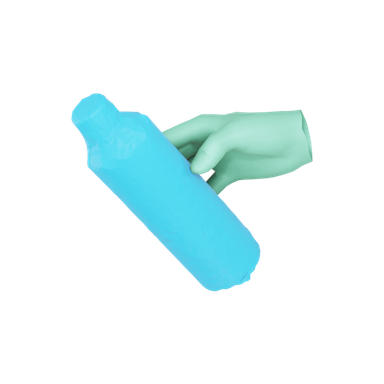}
\includegraphics[width=.90\linewidth]{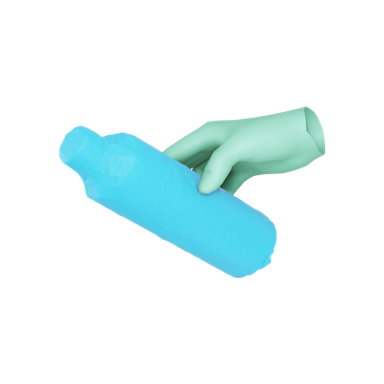}
\includegraphics[width=.90\linewidth]{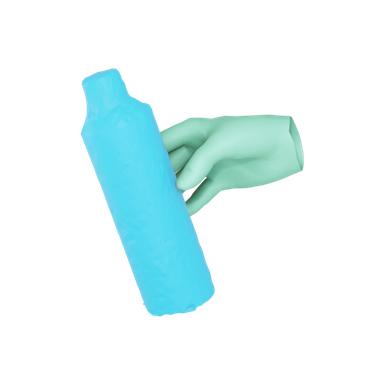}
\includegraphics[width=.90\linewidth]{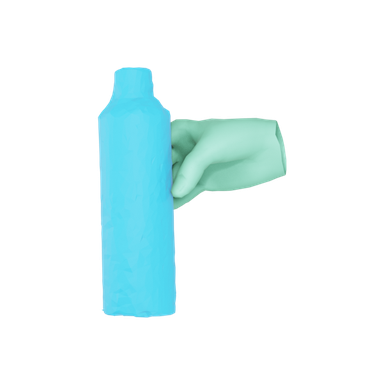}\\ \vspace{0mm} 
\includegraphics[width=.90\linewidth]{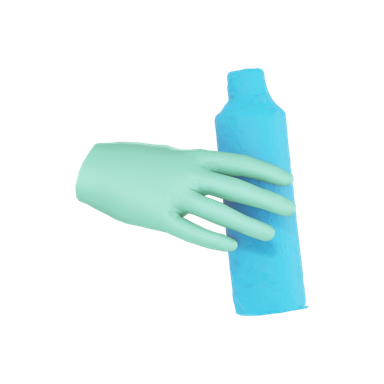}
\includegraphics[width=.90\linewidth]{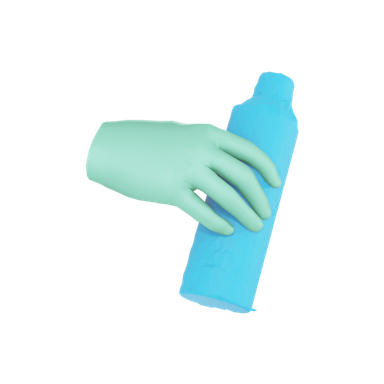}
\includegraphics[width=.90\linewidth]{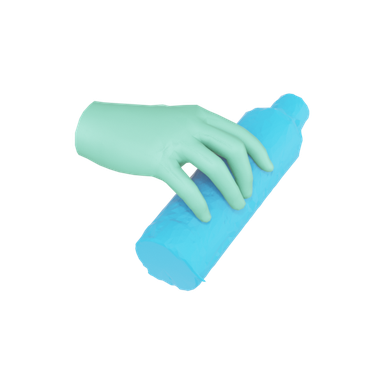}
\includegraphics[width=.90\linewidth]{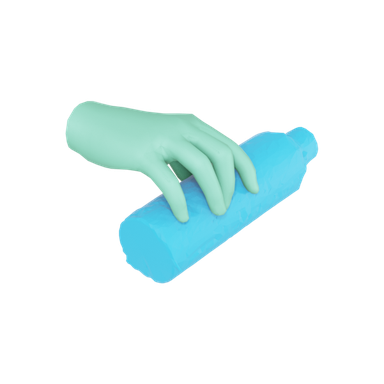}
\includegraphics[width=.90\linewidth]{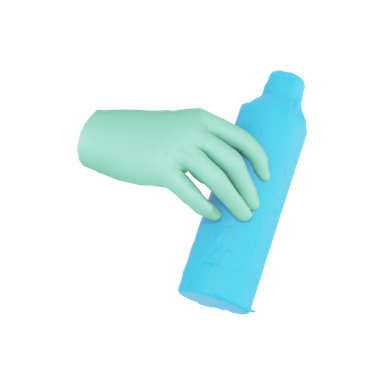}
\includegraphics[width=.90\linewidth]{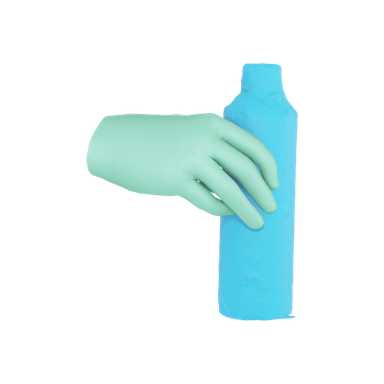} \\ \vspace{2mm} 

            \end{tabular}}
            
    \vspace{0.35cm}
    
    \resizebox{\textwidth}{!}{
            \begin{tabular}{cccccc} 
\includegraphics[width=.90\linewidth]{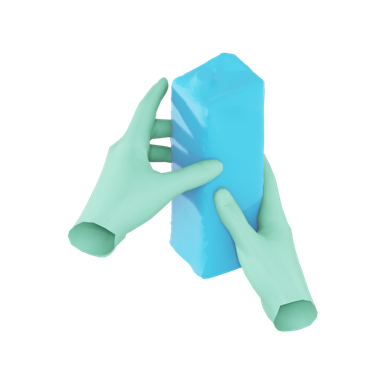}
\includegraphics[width=.90\linewidth]{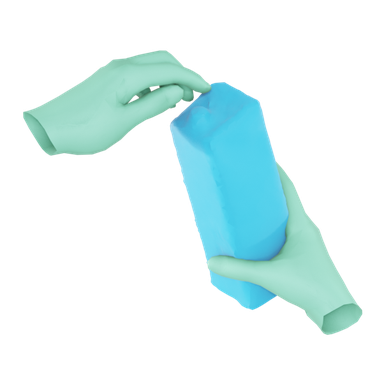}
\includegraphics[width=.90\linewidth]{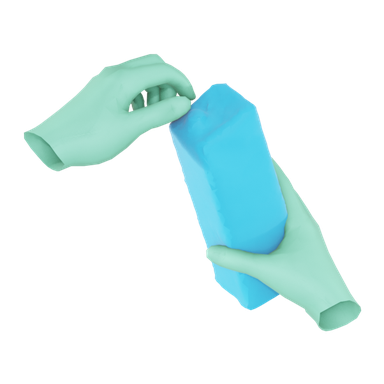}
\includegraphics[width=.90\linewidth]{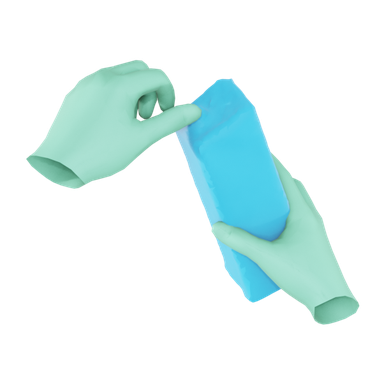}
\includegraphics[width=.90\linewidth]{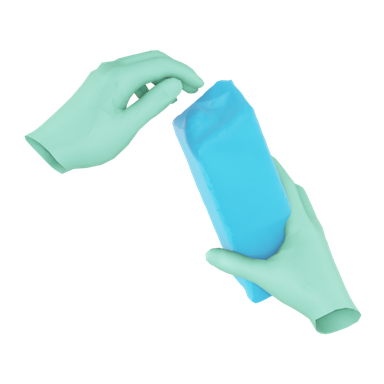}
\includegraphics[width=.90\linewidth]{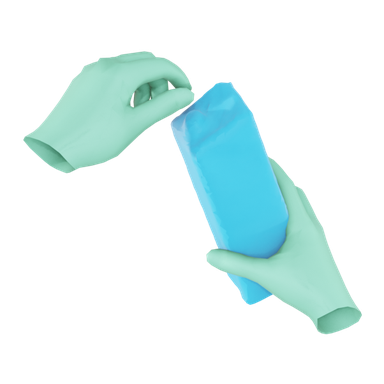} \\ \vspace{0mm} 
\includegraphics[width=.90\linewidth]{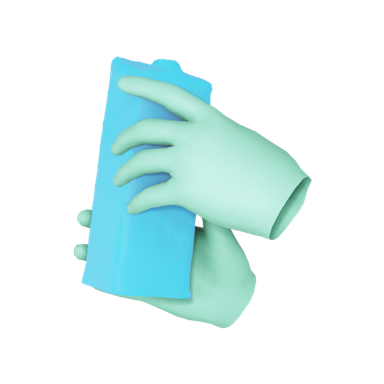}
\includegraphics[width=.90\linewidth]{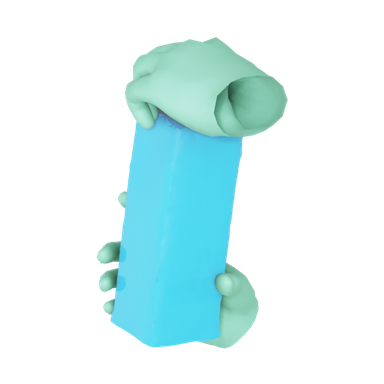}
\includegraphics[width=.90\linewidth]{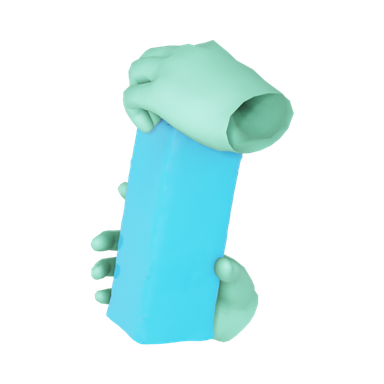}
\includegraphics[width=.90\linewidth]{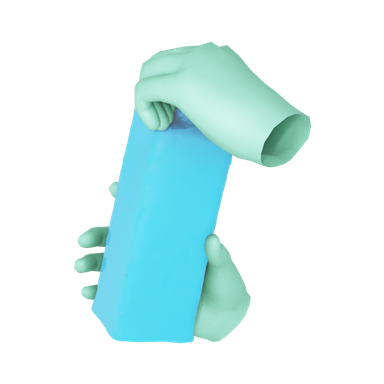}
\includegraphics[width=.90\linewidth]{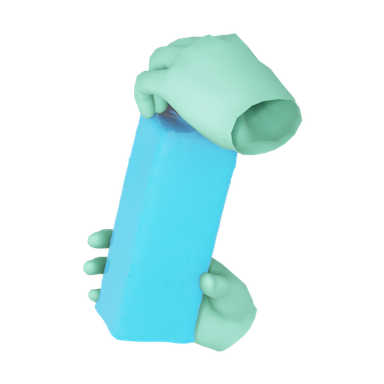}
\includegraphics[width=.90\linewidth]{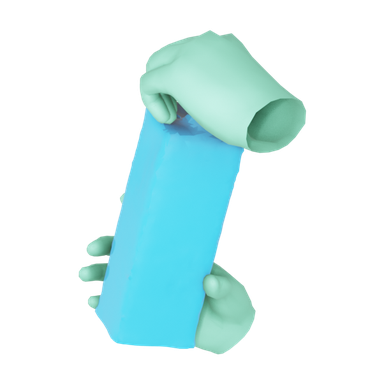}\\ \vspace{0mm} 
\includegraphics[width=.90\linewidth]{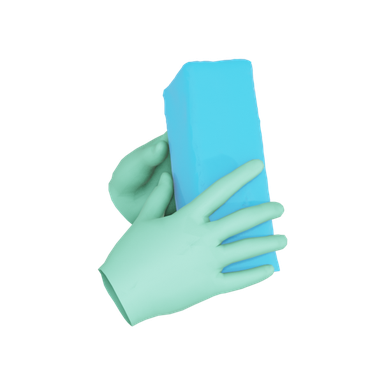}
\includegraphics[width=.90\linewidth]{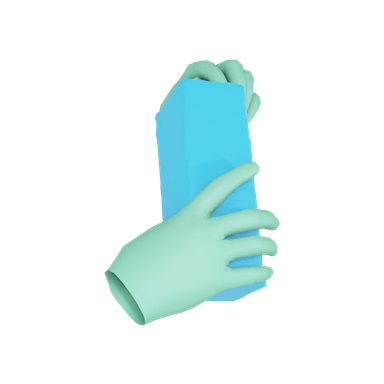}
\includegraphics[width=.90\linewidth]{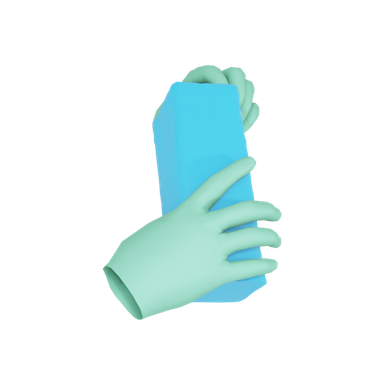}
\includegraphics[width=.90\linewidth]{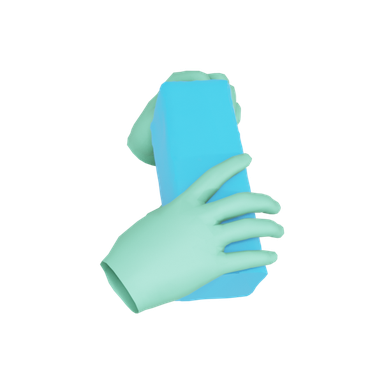}
\includegraphics[width=.90\linewidth]{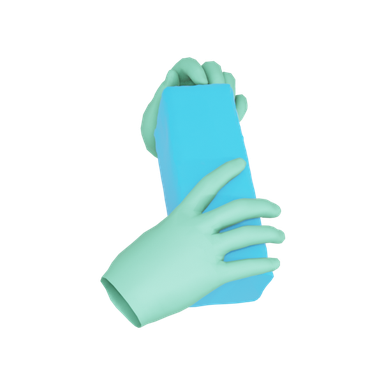}
\includegraphics[width=.90\linewidth]{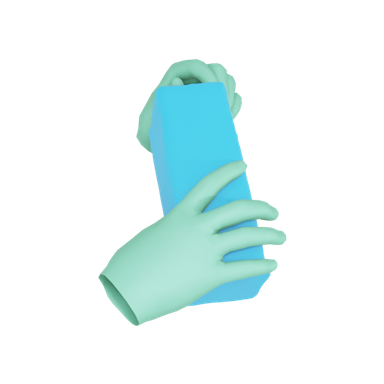} \\ \vspace{2mm} 

            \end{tabular}}


    \caption{Further examples of hand-object interaction trajectories generated by our method, as viewed from different perspectives. Video files with further visualizations are also available as part of the Supp.~Mat.}
    \label{fig:supp_rollouts}
\end{figure}

\begin{figure}[H]
    \centering
  \renewcommand{\arraystretch}{0}
	
    \resizebox{\textwidth}{!}{
            \begin{tabular}{ccc} 
\includegraphics[width=.85\linewidth]{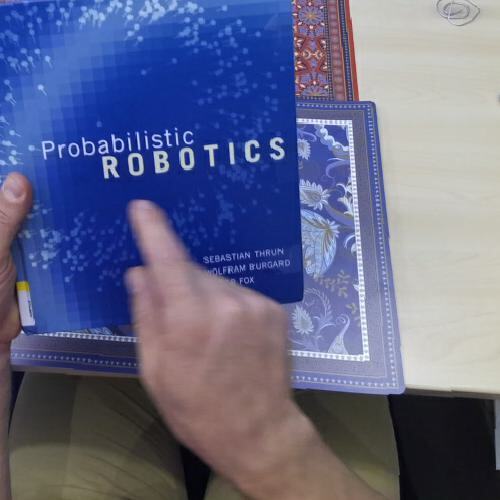} \hspace{0.2cm}
\includegraphics[width=.85\linewidth]{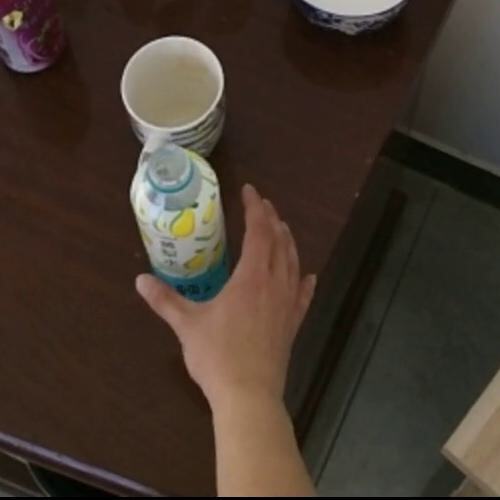} \hspace{0.2cm}
\includegraphics[width=.85\linewidth]{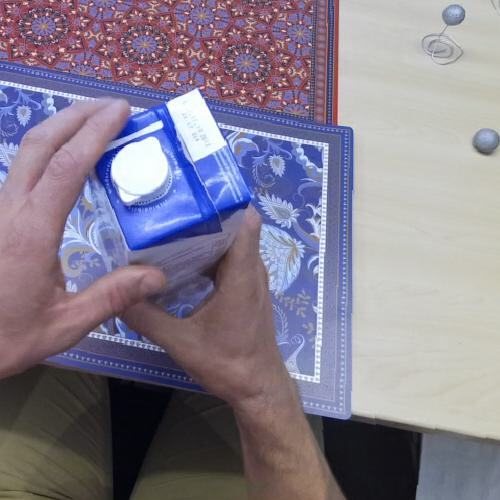}
            \end{tabular}}
    \caption{Conditioning images used to generate the trajectories in \autoref{fig:supp_rollouts} (top to bottom). The associated action labels are \textit{read book}, \textit{pour bottle}, \textit{open milk}.}
    \label{fig:supp_rollout_conditionings}
\end{figure}

We further provide a comparison of different variants of our proposed mesh retrieval scheme in \autoref{tab:mesh_retrieval_quant}. Specifically, we investigate the Chamfer distance to the ground-truth mesh when

(a) selecting a random mesh from the same category \textit{(Random)}, (b) selecting the mesh most similar based on visual features from the whole image \textit{(Image)}, as well as (c) selecting the mesh most similar based on visual features from a wrist-centered patch capturing the object \textit{(Object)}. The best performance when using setting (c) confirms the disambiguating property of retrieval based on visual features from a wrist-centered region closely capturing the hand-object interaction.

\begin{table}[htbp]
\footnotesize
\centering
\caption{Chamfer distance to ground-truth mesh, in centimeters, for different query types used to retrieve meshes from the HOI4D training set.}
\begin{tabular}{l c}
\toprule
{Setup} & {CD (cm) $\downarrow$} \\
\midrule
Random & $1.955\pm0.066$ \\
Image  & $1.718\pm0.000$\\
Object & $\textbf{1.618}\pm 0.000$ \\
\bottomrule
\end{tabular}
\vspace{0.2cm}
\label{tab:mesh_retrieval_quant}
\end{table}

\section{Further Qualitative Examples}

We provide additional visualizations of hand-object interaction trajectories generated by our method in \autoref{fig:supp_rollouts}. The respective conditioning images are provided in \autoref{fig:supp_rollout_conditionings}. Please also consult the video files attached to this supplementary material for further and richer qualitative results.

\section{Dataset Split}

{
\small
\begin{table}[h]
\begin{tabular}{l|l|l|l|llllllllllllllllll}
\cline{1-5}
\multicolumn{1}{c|}{{Action}}                 & \multicolumn{1}{c|}{{Original Task(s)}}                                & \multicolumn{1}{c|}{{Object Categories}}                                                                                                 & \multicolumn{1}{c|}{{\begin{tabular}[c]{@{}c@{}}Instances in\\ Training Set\end{tabular}}} & \multicolumn{1}{c}{{\begin{tabular}[c]{@{}c@{}}Instances in\\ Test Set\end{tabular}}} &  &  &  &  &  &  &  &  &  &  &  &  &  &  &  &  &  \\ \cline{1-5}
bind                                                      & bind the paper                                                                & stapler                                                                                                                                  & 38                                                                                             & 11                                                                                            &  &  &  &  &  &  &  &  &  &  &  &  &  &  &  &  &  \\ \cline{1-5}
clamp                                                     & clamp something                                                               & pliers                                                                                                                                   & 27                                                                                             & 9                                                                                             &  &  &  &  &  &  &  &  &  &  &  &  &  &  &  &  &  \\ \cline{1-5}
cut                                                       & cut something                                                                 & scissors                                                                                                                                 & 40                                                                                             & 12                                                                                            &  &  &  &  &  &  &  &  &  &  &  &  &  &  &  &  &  \\ \cline{1-5}
fill                                                      & fill with water by a kettle                                                   & mug                                                                                                                                      & 24                                                                                              & 7                                                                                             &  &  &  &  &  &  &  &  &  &  &  &  &  &  &  &  &  \\ \cline{1-5}
\multirow{1}{*}{open/close}                               & open and close the display                                                    & laptop                                                                                                                                   & 45                                                                                             & 13   \\ \cline{1-5}
\multirow{4}{*}{pour}                                     & pour all the water into a mug                                                 & bottle                                                                                                                                   & 30                                                                                             & 8                                                                                             &  &  &  &  &  &  &  &  &  &  &  &  &  &  &  &  &  \\ 
                                                          & pour water into a mug                                                         & kettle                                                                                                                                   & 40                                                                                             & 12                                                                                            &  &  &  &  &  &  &  &  &  &  &  &  &  &  &  &  &  \\ 
                                                          & pour water into another mug                                                   & mug                                                                                                                                      & 24                                                                                             & 7                                                                                             &  &  &  &  &  &  &  &  &  &  &  &  &  &  &  &  &  \\ 
                                                          & pour water into another bucket                                                & bucket                                                                                                                                   & 38                                                                                             & 11                                                                                            &  &  &  &  &  &  &  &  &  &  &  &  &  &  &  &  &  \\ \cline{1-5}
slice                                                     & cut apple                                                                     & knife                                                                                                                                    & 34                                                                                             & 9                                                                                            &  &  &  &  &  &  &  &  &  &  &  &  &  &  &  &  &  \\ \cline{1-5}
\end{tabular}\\[5pt]
\caption{\label{tab:hoi4d_split} \textbf{Actions for the HOI4D instance split.} Our proposed action-centric \textit{instance split} for the HOI4D dataset, assigning each instance of a class to either the train or the test set for each object and action. In this work, we train and evaluate on the objects \textit{bottle, bucket, kettle, knife, laptop, mug, pliers, scissors, stapler} from the HOI4D dataset.}
\end{table}
}

We list the \textit{actions} derived from the original tasks propose by HOI4D, as well as the corresponding instance counts per action and object category in the training and test set of our HOI4D instance set, in table \autoref{tab:hoi4d_split}.

For H2O, we follow the official action split provided by the authors.

\section{Hand Detection}

\begin{table}[]
\footnotesize
\centering
\caption{Detection rates of different approaches to locating the human hand in the input image. The Florence-2 \cite{Florence2} + SAM-2 \cite{SAM2} based approach detects hands in all images, yet struggles with correctly inferring hand sides.}
\begin{tabular}{ccccccc}
\hline
\multicolumn{1}{c|}{\multirow{3}{*}{Approach}} & \multicolumn{2}{c|}{\rule{0pt}{10pt}HOI4D}                                          & \multicolumn{4}{c}{H2O}                                                                                              \\
\multicolumn{1}{c|}{}                          & \multicolumn{1}{c|}{Training}    & \multicolumn{1}{c|}{Validation}  & \multicolumn{2}{c|}{Training}                                       & \multicolumn{2}{c}{Validation}                 \\
\multicolumn{1}{c|}{}                          & \multicolumn{1}{c|}{Det. Rate R} & \multicolumn{1}{l|}{Det. Rate R} & \multicolumn{1}{c|}{Det. Rate L} & \multicolumn{1}{c|}{Det. Rate R} & \multicolumn{1}{l|}{Det. Rate L} & Det. Rate R \\ \hline 
\multicolumn{1}{l}{\rule{0pt}{10pt}HaMeR \cite{HaMeR}}                      & \multicolumn{1}{c}{$95.91\%$}             & $92.17\%$      & \multicolumn{1}{c}{$93.31\%$}             &                        $94.37\%$   &      $98.36\%$                &      $100.00\%$       \\
\multicolumn{1}{l}{\begin{tabular}[c]{@{}l@{}}Florence-2 \cite{Florence2}\end{tabular}}         & \multicolumn{1}{c}{$100.00\%$}             &   $100.00\%$                               & \multicolumn{1}{c}{$100.00\%$}             &    $100.00\%$                       &  $100.00\%$      &   $100.00\%$   \\ \hline
\end{tabular}
\label{tab:wrist_ablation}

\end{table}

Detecting the human hand in the input image, as we wish to do so as to crop to the vicinity of the manipulated object, is prone to various types of errors, including failure to detect the hand (false negative) as well as incorrectly detecting the hand (false positive; e.g. detecting a hand of the wrong side).

We report the \textit{detection rates}, i.e. the percentage of frames in which a hand is found, of two different approaches to detecting the hand in \autoref{tab:wrist_ablation}. Specifically, we consider the idea of using the hand pose reconstruction method HaMeR \cite{HaMeR} to detect the wrist keypoint of the hand in the image. Note that this method provides an estimated hand side along with the hand keypoints. We further investigate the use of obtaining a hand bounding box by grounding with Florence-2 \cite{Florence2}, conditioned with the prompts ``human hand" (HOI4D) and ``right hand" resp.~``left hand" (H2O). We further investigate how well each approach distinguishes hand sides in \autoref{tab:h2o_hand_correctness_stats}.

\begin{table}[t!]
\footnotesize
\centering
\caption{Statistics of true positive (TP), false positive (FP) and false negative (FN) right hand detections on the H2O validation set (122 samples) for different approaches. The HaMeR \cite{HaMeR}-based method is reliable in its prediction of the hand side, while the Florence-2\cite{Florence2}--based approach cannot distinguish left from right hands, leading to a high number of false positives and false negatives.}
\begin{tabular}{cccc}
\hline
Approach     & TP  & FP & FN \\ \hline
\rule{0pt}{8pt}HaMeR \cite{HaMeR}      & 121 & 1  & 1  \\ 
Florence-2 \cite{Florence2} & 77  & 88 & 45 \\ \hline
\end{tabular}
\label{tab:h2o_hand_correctness_stats}
\end{table}

As evident from the results, Florence-2 achieves full recall on both training and validation splits of both datasets, while HaMeR consistently detects hands in about 90\%-100\% of all images. On HOI4D, where we consider interactions involving only the right hand, we can achieve a perfect detection rate by using Florence-2, or alternatively using HaMeR and then Florence-2 on images where HaMeR did not give hand detections. On the bimanual H2O, using Florence-2 will lead to more uncontrolled results, as the model frequently confuses left and right hands. HaMeR is able to detect all right hands, and almost all left hands on H2O's validation set. To always know to which region to crop, we hence opt to use the region surrounding the right hand based on HaMeR's wrist position prediction when performing inference on H2O.

\section{Hyperparameters}

We use $T=1000$ denoising steps for all diffusion-based models. For our guidance term, we experimented with different schedules and magnitudes of the guidance scale and empirically found a constant scale of 7, applied after the initial 100 denoising steps, to lead to the best results.

All models are trained for 250 epochs. Consistent with the original implementations, ReMoDiffuse \cite{ReMoDiffuse} and MotionDiffuse \cite{MotionDiffuse}-based models are trained using the Adam \cite{Adam} optimizer with a learning rate of $2\cdot10^{-4}$, as well as a batch size of 128. MDM \cite{MDM}-based models use a learning rate of $1\cdot10^{-4}$ and a batch size of 64.

{
\small
\bibliography{egbib}

\begin{thebibliography}{68}
\providecommand{\natexlab}[1]{#1}
\providecommand{\url}[1]{\texttt{#1}}
\expandafter\ifx\csname urlstyle\endcsname\relax
  \providecommand{\doi}[1]{doi: #1}\else
  \providecommand{\doi}{doi: \begingroup \urlstyle{rm}\Url}\fi

\bibitem[Aksan et~al.(2021)Aksan, Kaufmann, Cao, and Hilliges]{aksan2021spatio}
Emre Aksan, Manuel Kaufmann, Peng Cao, and Otmar Hilliges.
\newblock A spatio-temporal transformer for 3d human motion prediction.
\newblock In \emph{International Conference on 3D Vision (3DV)}, 2021.

\bibitem[Aliakbarian et~al.(2020)Aliakbarian, Saleh, Salzmann, Petersson, and Gould]{mix-and-match-perturbation}
Sadegh Aliakbarian, Fatemeh~Sadat Saleh, Mathieu Salzmann, Lars Petersson, and Stephen Gould.
\newblock A stochastic conditioning scheme for diverse human motion prediction.
\newblock In \emph{ICCV}, 2020.

\bibitem[Cai et~al.(2021)Cai, Wang, Zhu, Cham, Cai, Yuan, Liu, Zheng, Yan, Ding, et~al.]{cai2021unified}
Yujun Cai, Yiwei Wang, Yiheng Zhu, Tat-Jen Cham, Jianfei Cai, Junsong Yuan, Jun Liu, Chuanxia Zheng, Sijie Yan, Henghui Ding, et~al.
\newblock A unified 3d human motion synthesis model via conditional variational auto-encoder.
\newblock In \emph{CVPR}, 2021.

\bibitem[Cao et~al.(2017)Cao, Simon, Wei, and Sheikh]{OpenPose}
Zhe Cao, Tomas Simon, Shih-En Wei, and Yaser Sheikh.
\newblock {Realtime Multi-Person 2D Pose Estimation Using Part Affinity Fields}.
\newblock In \emph{{Proceedings of the IEEE Conference on Computer Vision and Pattern Recognition}}, pages 7291--7299, 2017.

\bibitem[Cao et~al.(2020)Cao, Gao, Mangalam, Cai, Vo, and Malik]{caoHMP2020}
Zhe Cao, Hang Gao, Karttikeya Mangalam, Qizhi Cai, Minh Vo, and Jitendra Malik.
\newblock Long-term human motion prediction with scene context.
\newblock In \emph{ECCV}, 2020.

\bibitem[Cha et~al.(2024)Cha, Kim, Yoon, and Baek]{Text2HOI}
Junuk Cha, Jihyeon Kim, Jae~Shin Yoon, and Seungryul Baek.
\newblock {Text2HOI: Text-guided 3D Motion Generation for Hand-Object Interaction}.
\newblock In \emph{CVPR}, 2024.

\bibitem[Chen et~al.(2023)Chen, Jiang, Liu, Huang, Fu, Chen, and Yu]{chen2023executing}
Xin Chen, Biao Jiang, Wen Liu, Zilong Huang, Bin Fu, Tao Chen, and Gang Yu.
\newblock Executing your commands via motion diffusion in latent space.
\newblock In \emph{CVPR}, 2023.

\bibitem[Christen et~al.(2022)Christen, Kocabas, Aksan, Hwangbo, Song, and Hilliges]{christen2022dgrasp}
Sammy Christen, Muhammed Kocabas, Emre Aksan, Jemin Hwangbo, Jie Song, and Otmar Hilliges.
\newblock D-grasp: Physically plausible dynamic grasp synthesis for hand-object interactions.
\newblock In \emph{CVPR}, 2022.

\bibitem[Christen et~al.(2024)Christen, Hampali, Sener, Remelli, Hodan, Sauser, Ma, and Tekin]{christen2024diffh2o}
Sammy Christen, Shreyas Hampali, Fadime Sener, Edoardo Remelli, Tomas Hodan, Eric Sauser, Shugao Ma, and Bugra Tekin.
\newblock Diffh2o: Diffusion-based synthesis of hand-object interactions from textual descriptions.
\newblock In \emph{SIGGRAPH Asia 2024 Conference Papers}, 2024.

\bibitem[Dabral et~al.(2023)Dabral, Mughal, Golyanik, and Theobalt]{dabral2022mofusion}
Rishabh Dabral, Muhammad~Hamza Mughal, Vladislav Golyanik, and Christian Theobalt.
\newblock {MoFusion: A Framework for Denoising-Diffusion-based Motion Synthesis}.
\newblock In \emph{Computer Vision and Pattern Recognition (CVPR)}, 2023.

\bibitem[Darkhalil et~al.(2022)Darkhalil, Shan, Zhu, Ma, Kar, Higgins, Fidler, Fouhey, and Damen]{EKVisor}
Ahmad Darkhalil, Dandan Shan, Bin Zhu, Jian Ma, Amlan Kar, Richard Higgins, Sanja Fidler, David Fouhey, and Dima Damen.
\newblock {{EPIC-KITCHENS VISOR Benchmark}}: {{VIdeo Segmentations}} and {{Object Relations}}.
\newblock In \emph{Proceedings of the {{Neural Information Processing Systems}} ({{NeurIPS}}) {{Track}} on {{Datasets}} and {{Benchmarks}}}, 2022.

\bibitem[Dhariwal and Nichol(2021{\natexlab{a}})]{ClassifierGuidance}
Prafulla Dhariwal and Alexander Nichol.
\newblock {Diffusion Models Beat Gans on Image Synthesis}.
\newblock \emph{Advances in neural information processing systems}, 34:\penalty0 8780--8794, 2021{\natexlab{a}}.

\bibitem[Dhariwal and Nichol(2021{\natexlab{b}})]{dhariwal2021diffusion}
Prafulla Dhariwal and Alexander Nichol.
\newblock Diffusion models beat gans on image synthesis.
\newblock In \emph{NeurIPS}, 2021{\natexlab{b}}.

\bibitem[Diomataris et~al.(2024)Diomataris, Athanasiou, Taheri, Wang, Hilliges, and Black]{diomataris2024wandr}
Markos Diomataris, Nikos Athanasiou, Omid Taheri, Xi~Wang, Otmar Hilliges, and Michael~J. Black.
\newblock Wandr: Intention-guided human motion generation.
\newblock In \emph{CVPR}, 2024.

\bibitem[Epstein et~al.(2023)Epstein, Jabri, Poole, Efros, and Holynski]{epstein2023selfguidance}
Dave Epstein, Allan Jabri, Ben Poole, Alexei~A. Efros, and Aleksander Holynski.
\newblock Diffusion self-guidance for controllable image generation.
\newblock 2023.

\bibitem[Fan et~al.(2024)Fan, Parelli, Kadoglou, Kocabas, Chen, Black, and Hilliges]{fan2024hold}
Zicong Fan, Maria Parelli, Maria~Eleni Kadoglou, Muhammed Kocabas, Xu~Chen, Michael~J Black, and Otmar Hilliges.
\newblock {HOLD}: {Category-agnostic 3D Reconstruction of Interacting Hands and Objects from Video}.
\newblock 2024.

\bibitem[Ghosh et~al.(2023)Ghosh, Dabral, Golyanik, Theobalt, and Slusallek]{ghosh2022imos}
Anindita Ghosh, Rishabh Dabral, Vladislav Golyanik, Christian Theobalt, and Philipp Slusallek.
\newblock Imos: Intent-driven full-body motion synthesis for human-object interactions.
\newblock In \emph{Eurographics}, 2023.

\bibitem[Guo et~al.(2020{\natexlab{a}})Guo, Zuo, Wang, Zou, Sun, Deng, Gong, and Cheng]{Action2Motion}
Chuan Guo, Xinxin Zuo, Sen Wang, Shihao Zou, Qingyao Sun, Annan Deng, Minglun Gong, and Li~Cheng.
\newblock Action2motion: {Conditioned Generation of 3d Human Motions}.
\newblock In \emph{Proceedings of the 28th {{ACM International Conference}} on {{Multimedia}}}, pages 2021--2029, 2020{\natexlab{a}}.

\bibitem[Guo et~al.(2020{\natexlab{b}})Guo, Zuo, Wang, Zou, Sun, Deng, Gong, and Cheng]{guo2020action2motion}
Chuan Guo, Xinxin Zuo, Sen Wang, Shihao Zou, Qingyao Sun, Annan Deng, Minglun Gong, and Li~Cheng.
\newblock {Action2motion: Conditioned Generation of 3D Human Motions}.
\newblock In \emph{Proceedings of the 28th ACM International Conference on Multimedia}, pages 2021--2029, 2020{\natexlab{b}}.

\bibitem[Guo et~al.(2022{\natexlab{a}})Guo, Zou, Zuo, Wang, Ji, Li, and Cheng]{t2m}
Chuan Guo, Shihao Zou, Xinxin Zuo, Sen Wang, Wei Ji, Xingyu Li, and Li~Cheng.
\newblock {Generating Diverse and Natural 3D Human Motions From Text}.
\newblock In \emph{{Proceedings of the IEEE/CVF Conference on Computer Vision and Pattern Recognition (CVPR)}}, pages 5152--5161, June 2022{\natexlab{a}}.

\bibitem[Guo et~al.(2022{\natexlab{b}})Guo, Zuo, Wang, and Cheng]{tm2t}
Chuan Guo, Xinxin Zuo, Sen Wang, and Li~Cheng.
\newblock {TM2T: Stochastic and Tokenized Modeling for the Reciprocal Generation of 3D Human Motions and Texts}.
\newblock In \emph{ECCV}, 2022{\natexlab{b}}.

\bibitem[Hasson et~al.(2019)Hasson, Varol, Tzionas, Kalevatykh, Black, Laptev, and Schmid]{Hasson19LearningJointReconstruction}
Yana Hasson, Gul Varol, Dimitrios Tzionas, Igor Kalevatykh, Michael~J Black, Ivan Laptev, and Cordelia Schmid.
\newblock Learning {{Joint Reconstruction}} of {{Hands}} and {{Manipulated Objects}}.
\newblock In \emph{Proceedings of the {{IEEE}}/{{CVF Conference}} on {{Computer Vision}} and {{Pattern Recognition}}}, pages 11807--11816, 2019.

\bibitem[Heusel et~al.(2017)Heusel, Ramsauer, Unterthiner, Nessler, and Hochreiter]{FID}
Martin Heusel, Hubert Ramsauer, Thomas Unterthiner, Bernhard Nessler, and Sepp Hochreiter.
\newblock {GANs Trained by a Two Time-Scale Update Rule Converge to a Local Nash Equilibrium}.
\newblock \emph{{Advances in Neural Information Processing Systems}}, 30, 2017.

\bibitem[Ho et~al.(2020)Ho, Jain, and Abbeel]{ho2020ddpm}
Jonathan Ho, Ajay Jain, and Pieter Abbeel.
\newblock Denoising diffusion probabilistic models.
\newblock In \emph{NeurIPS}, 2020.

\bibitem[Jiang et~al.(2024)Jiang, Chen, Liu, Yu, Yu, and Chen]{jiang2024motiongpt}
Biao Jiang, Xin Chen, Wen Liu, Jingyi Yu, Gang Yu, and Tao Chen.
\newblock Motiongpt: Human motion as a foreign language.
\newblock \emph{NeurIPS}, 36, 2024.

\bibitem[Jiang et~al.(2021)Jiang, Liu, Wang, and Wang]{jiang2021graspTTA}
Hanwen Jiang, Shaowei Liu, Jiashun Wang, and Xiaolong Wang.
\newblock Hand-object contact consistency reasoning for human grasps generation.
\newblock In \emph{ICCV}, 2021.

\bibitem[Karunratanakul et~al.(2023)Karunratanakul, Preechakul, Suwajanakorn, and Tang]{karunratanakul2023gmd}
Korrawe Karunratanakul, Konpat Preechakul, Supasorn Suwajanakorn, and Siyu Tang.
\newblock {GMD: Controllable Human Motion Synthesis via Guided Diffusion Models}.
\newblock In \emph{CVPR}, 2023.

\bibitem[Kim et~al.(2023)Kim, Kim, and Choi]{kim2023flame}
Jihoon Kim, Jiseob Kim, and Sungjoon Choi.
\newblock {FLAME: Free-form Language-based Motion Synthesis \& Editing}, 2023.

\bibitem[Kingma(2014)]{Adam}
Diederik~P Kingma.
\newblock {Adam: A Method for Stochastic Optimization}.
\newblock \emph{arXiv preprint arXiv:1412.6980}, 2014.

\bibitem[Kwon et~al.(2021)Kwon, Tekin, St\"uhmer, Bogo, and Pollefeys]{H2O}
Taein Kwon, Bugra Tekin, Jan St\"uhmer, Federica Bogo, and Marc Pollefeys.
\newblock {H2O: Two Hands Manipulating Objects for First Person Interaction Recognition}.
\newblock In \emph{{Proceedings of the IEEE/CVF International Conference on Computer Vision (ICCV)}}, pages 10138--10148, October 2021.

\bibitem[Liang et~al.(2024)Liang, Zhang, Li, Yu, and Xu]{liang2024intergen}
Han Liang, Wenqian Zhang, Wenxuan Li, Jingyi Yu, and Lan Xu.
\newblock Intergen: Diffusion-based multi-human motion generation under complex interactions.
\newblock \emph{IJCV}, 2024.

\bibitem[Ling et~al.(2020)Ling, Zinno, Cheng, and Panne]{ling2020character}
Hung~Yu Ling, Fabio Zinno, George Cheng, and Michiel Van~De Panne.
\newblock Character controllers using motion vaes.
\newblock \emph{ACM TOG}, 2020.

\bibitem[Liu et~al.(2022)Liu, Liu, Jiang, Lyu, Wan, Shen, Liang, Fu, Wang, and Yi]{HOI4D}
Yunze Liu, Yun Liu, Che Jiang, Kangbo Lyu, Weikang Wan, Hao Shen, Boqiang Liang, Zhoujie Fu, He~Wang, and Li~Yi.
\newblock {{HOI4D}}: {{A 4D}} egocentric dataset for category-level human-object interaction.
\newblock In \emph{Proceedings of the {{IEEE}}/{{CVF}} Conference on Computer Vision and Pattern Recognition}, pages 21013--21022, 2022.

\bibitem[Pavlakos et~al.(2024)Pavlakos, Shan, Radosavovic, Kanazawa, Fouhey, and Malik]{HaMeR}
Georgios Pavlakos, Dandan Shan, Ilija Radosavovic, Angjoo Kanazawa, David Fouhey, and Jitendra Malik.
\newblock {Reconstructing Hands in 3D with Transformers}.
\newblock In \emph{CVPR}, 2024.

\bibitem[Petrovich et~al.(2022)Petrovich, Black, and Varol]{petrovich2022temos}
Mathis Petrovich, Michael~J Black, and G{\"u}l Varol.
\newblock Temos: Generating diverse human motions from textual descriptions.
\newblock In \emph{ECCV}, 2022.

\bibitem[Radford et~al.(2021)Radford, Kim, Hallacy, Ramesh, Goh, Agarwal, Sastry, Askell, Mishkin, Clark, et~al.]{CLIP}
Alec Radford, Jong~Wook Kim, Chris Hallacy, Aditya Ramesh, Gabriel Goh, Sandhini Agarwal, Girish Sastry, Amanda Askell, Pamela Mishkin, Jack Clark, et~al.
\newblock Learning {{Transferable Visual Models From Natural Language Supervision}}.
\newblock In \emph{International {{Conference}} on {{Machine Learning}}}, pages 8748--8763. {PMLR}, 2021.

\bibitem[Ravi et~al.(2024)Ravi, Gabeur, Hu, Hu, Ryali, Ma, Khedr, R{\"a}dle, Rolland, Gustafson, Mintun, Pan, Alwala, Carion, Wu, Girshick, Doll{\'a}r, and Feichtenhofer]{SAM2}
Nikhila Ravi, Valentin Gabeur, Yuan-Ting Hu, Ronghang Hu, Chaitanya Ryali, Tengyu Ma, Haitham Khedr, Roman R{\"a}dle, Chloe Rolland, Laura Gustafson, Eric Mintun, Junting Pan, Kalyan~Vasudev Alwala, Nicolas Carion, Chao-Yuan Wu, Ross Girshick, Piotr Doll{\'a}r, and Christoph Feichtenhofer.
\newblock {SAM 2: Segment Anything in Images and Videos}.
\newblock \emph{arXiv preprint arXiv:2408.00714}, 2024.
\newblock URL \url{https://arxiv.org/abs/2408.00714}.

\bibitem[Rombach et~al.(2021)Rombach, Blattmann, Lorenz, Esser, and Ommer]{StableDiffusion}
Robin Rombach, Andreas Blattmann, Dominik Lorenz, Patrick Esser, and Björn Ommer.
\newblock High-{{Resolution Image Synthesis}} with {{Latent Diffusion Models}}, 2021.

\bibitem[Romero et~al.(2017-11)Romero, Tzionas, and Black]{MANO}
Javier Romero, Dimitrios Tzionas, and Michael~J. Black.
\newblock Embodied {{Hands}}: {{Modeling}} and {{Capturing Hands}} and {{Bodies Together}}.
\newblock 36\penalty0 (6), 2017-11.

\bibitem[Shafir et~al.(2024)Shafir, Tevet, Kapon, and Bermano]{shafir2024priormdm}
Yoni Shafir, Guy Tevet, Roy Kapon, and Amit~Haim Bermano.
\newblock Human motion diffusion as a generative prior.
\newblock In \emph{ICLR}, 2024.

\bibitem[Shimada et~al.(2024)Shimada, Mueller, Bednarik, Doosti, Bickel, Tang, Golyanik, Taylor, Theobalt, and Beeler]{MACS2024}
Soshi Shimada, Franziska Mueller, Jan Bednarik, Bardia Doosti, Bernd Bickel, Danhang Tang, Vladislav Golyanik, Jonathan Taylor, Christian Theobalt, and Thabo Beeler.
\newblock Macs: Mass conditioned 3d hand and object motion synthesis.
\newblock In \emph{International Conference on 3D Vision (3DV)}, 2024.

\bibitem[Shu et~al.(2022)Shu, Zhang, Qi, Liu, and Tang]{Xiangbo2022recurrent}
Xiangbo Shu, Liyan Zhang, Guo-Jun Qi, Wei Liu, and Jinhui Tang.
\newblock { Spatiotemporal Co-Attention Recurrent Neural Networks for Human-Skeleton Motion Prediction }.
\newblock \emph{IEEE TPAMI}, 44\penalty0 (06):\penalty0 3300--3315, June 2022.
\newblock ISSN 1939-3539.

\bibitem[Tevet et~al.(2022{\natexlab{a}})Tevet, Gordon, Hertz, Bermano, and Cohen-Or]{tevet2022motionclip}
Guy Tevet, Brian Gordon, Amir Hertz, Amit~H Bermano, and Daniel Cohen-Or.
\newblock Motionclip: Exposing human motion generation to clip space.
\newblock In \emph{ECCV}, 2022{\natexlab{a}}.

\bibitem[Tevet et~al.(2022{\natexlab{b}})Tevet, Raab, Gordon, Shafir, Cohen-Or, and Bermano]{MDM}
Guy Tevet, Sigal Raab, Brian Gordon, Yonatan Shafir, Daniel Cohen-Or, and Amit~H Bermano.
\newblock Human {{Motion Diffusion Model}}.
\newblock 2022{\natexlab{b}}.

\bibitem[Tevet et~al.(2023{\natexlab{a}})Tevet, Raab, Gordon, Shafir, Cohen-or, and Bermano]{tevet2023human}
Guy Tevet, Sigal Raab, Brian Gordon, Yoni Shafir, Daniel Cohen-or, and Amit~Haim Bermano.
\newblock {Human Motion Diffusion Model}.
\newblock In \emph{The Eleventh International Conference on Learning Representations}, 2023{\natexlab{a}}.
\newblock URL \url{https://openreview.net/forum?id=SJ1kSyO2jwu}.

\bibitem[Tevet et~al.(2023{\natexlab{b}})Tevet, Raab, Gordon, Shafir, Cohen-or, and Bermano]{tevet2023mdm}
Guy Tevet, Sigal Raab, Brian Gordon, Yoni Shafir, Daniel Cohen-or, and Amit~Haim Bermano.
\newblock {Human Motion Diffusion Model}.
\newblock In \emph{ICLR}, 2023{\natexlab{b}}.

\bibitem[Vaswani et~al.(2017)Vaswani, Shazeer, Parmar, Uszkoreit, Jones, Gomez, Kaiser, and Polosukhin]{vaswani2017attention}
Ashish Vaswani, Noam Shazeer, Niki Parmar, Jakob Uszkoreit, Llion Jones, Aidan~N Gomez, \L~ukasz Kaiser, and Illia Polosukhin.
\newblock Attention is all you need.
\newblock In \emph{NeurIPS}, 2017.

\bibitem[Wang et~al.(2020{\natexlab{a}})Wang, Arti{\`e}res, Chen, and Denoyer]{wang2020adversarial}
Qi~Wang, Thierry Arti{\`e}res, Mickael Chen, and Ludovic Denoyer.
\newblock Adversarial learning for modeling human motion.
\newblock \emph{The Visual Computer}, 36\penalty0 (1):\penalty0 141--160, 2020{\natexlab{a}}.

\bibitem[Wang et~al.(2020{\natexlab{b}})Wang, Yu, Zhao, Zhang, Zhou, Yuan, and Chen]{wang2020learning}
Zhenyi Wang, Ping Yu, Yang Zhao, Ruiyi Zhang, Yufan Zhou, Junsong Yuan, and Changyou Chen.
\newblock Learning diverse stochastic human-action generators by learning smooth latent transitions.
\newblock In \emph{AAAI}, 2020{\natexlab{b}}.

\bibitem[Xiao et~al.(2024)Xiao, Wu, Xu, Dai, Hu, Lu, Zeng, Liu, and Yuan]{Florence2}
Bin Xiao, Haiping Wu, Weijian Xu, Xiyang Dai, Houdong Hu, Yumao Lu, Michael Zeng, Ce~Liu, and Lu~Yuan.
\newblock {Florence-2: Advancing a Unified Representation for a Variety of Vision Tasks}.
\newblock In \emph{{Proceedings of the IEEE/CVF Conference on Computer Vision and Pattern Recognition}}, pages 4818--4829, 2024.

\bibitem[Ye et~al.(2022)Ye, Gupta, and Tulsiani]{Ye22WhatsInYourHands}
Yufei Ye, Abhinav Gupta, and Shubham Tulsiani.
\newblock What’s in {{Your Hands}}? {{3D Reconstruction}} of {{Generic Objects}} in {{Hands}}.
\newblock In \emph{Proceedings of the {{IEEE}}/{{CVF Conference}} on {{Computer Vision}} and {{Pattern Recognition}}}, pages 3895--3905, 2022.

\bibitem[Ye et~al.(2023)Ye, Hebbar, Gupta, and Tulsiani]{ye2023vhoi}
Yufei Ye, Poorvi Hebbar, Abhinav Gupta, and Shubham Tulsiani.
\newblock {Diffusion-Guided Reconstruction of Everyday Hand-Object Interaction Clips}.
\newblock In \emph{ICCV}, 2023.

\bibitem[Yi et~al.(2023)Yi, Liang, Liu, Cao, Wen, Bolkart, Tao, and Black]{generatingHuman}
Hongwei Yi, Hualin Liang, Yifei Liu, Qiong Cao, Yandong Wen, Timo Bolkart, Dacheng Tao, and Michael~J. Black.
\newblock {Generating Holistic 3D Human Motion from Speech}, 2023.

\bibitem[Yi et~al.(2024)Yi, Thies, Black, Peng, and Rempe]{yi2024tesmo}
Hongwei Yi, Justus Thies, Michael~J. Black, Xue~Bin Peng, and Davis Rempe.
\newblock Generating human interaction motions in scenes with text control.
\newblock In \emph{ECCV}, 2024.

\bibitem[Yu et~al.(2020)Yu, Zhao, Li, Yuan, and Chen]{yu2020structure}
Ping Yu, Yang Zhao, Chunyuan Li, Junsong Yuan, and Changyou Chen.
\newblock Structure-aware human-action generation.
\newblock In \emph{ECCV}, 2020.

\bibitem[Zhai et~al.(2023)Zhai, Mustafa, Kolesnikov, and Beyer]{SigLIP}
Xiaohua Zhai, Basil Mustafa, Alexander Kolesnikov, and Lucas Beyer.
\newblock {Sigmoid Loss for Language Image Pre-training}.
\newblock In \emph{{Proceedings of the IEEE/CVF International Conference on Computer Vision}}, pages 11975--11986, 2023.

\bibitem[Zhang et~al.(2021)Zhang, Ye, Shiratori, and Komura]{manipnet}
He~Zhang, Yuting Ye, Takaaki Shiratori, and Taku Komura.
\newblock Manipnet: neural manipulation synthesis with a hand-object spatial representation.
\newblock \emph{ACM TOG}, 40\penalty0 (4), July 2021.
\newblock ISSN 0730-0301.

\bibitem[Zhang et~al.(2024{\natexlab{a}})Zhang, Christen, Fan, Hilliges, and Song]{zhang2024graspxl}
Hui Zhang, Sammy Christen, Zicong Fan, Otmar Hilliges, and Jie Song.
\newblock {GraspXL}: Generating grasping motions for diverse objects at scale.
\newblock In \emph{ECCV}, 2024{\natexlab{a}}.

\bibitem[Zhang et~al.(2024{\natexlab{b}})Zhang, Christen, Fan, Zheng, Hwangbo, Song, and Hilliges]{zhang2024artigrasp}
Hui Zhang, Sammy Christen, Zicong Fan, Luocheng Zheng, Jemin Hwangbo, Jie Song, and Otmar Hilliges.
\newblock {ArtiGrasp}: Physically plausible synthesis of bi-manual dexterous grasping and articulation.
\newblock In \emph{International Conference on 3D Vision (3DV)}, 2024{\natexlab{b}}.

\bibitem[Zhang et~al.(2024{\natexlab{c}})Zhang, Zhang, An, Li, Zhang, Hu, and Liu]{zhang2024manidext}
Jiajun Zhang, Yuxiang Zhang, Liang An, Mengcheng Li, Hongwen Zhang, Zonghai Hu, and Yebin Liu.
\newblock Manidext: Hand-object manipulation synthesis via continuous correspondence embeddings and residual-guided diffusion, 2024{\natexlab{c}}.
\newblock URL \url{https://arxiv.org/abs/2409.09300}.

\bibitem[Zhang et~al.(2023{\natexlab{a}})Zhang, Zhang, Cun, Huang, Zhang, Zhao, Lu, and Shen]{t2mgpt}
Jianrong Zhang, Yangsong Zhang, Xiaodong Cun, Shaoli Huang, Yong Zhang, Hongwei Zhao, Hongtao Lu, and Xi~Shen.
\newblock {T2M-GPT: Generating Human Motion from Textual Descriptions with Discrete Representations}, 2023{\natexlab{a}}.

\bibitem[Zhang et~al.(2022{\natexlab{a}})Zhang, Zhou, Stent, and Shi]{Zhang22EgoHOS}
Lingzhi Zhang, Shenghao Zhou, Simon Stent, and Jianbo Shi.
\newblock Fine-{{Grained Egocentric Hand-Object Segmentation}}: {{Dataset}}, {{Model}}, and {{Applications}}.
\newblock In \emph{European Conference on Computer Vision}, pages 127--145. {Springer}, 2022{\natexlab{a}}.

\bibitem[Zhang et~al.(2022{\natexlab{b}})Zhang, Cai, Pan, Hong, Guo, Yang, and Liu]{MotionDiffuse}
Mingyuan Zhang, Zhongang Cai, Liang Pan, Fangzhou Hong, Xinying Guo, Lei Yang, and Ziwei Liu.
\newblock {MotionDiffuse: Text-Driven Human Motion Generation With Diffusion Model}.
\newblock \emph{arXiv preprint arXiv:2208.15001}, 2022{\natexlab{b}}.

\bibitem[Zhang et~al.(2023{\natexlab{b}})Zhang, Guo, Pan, Cai, Hong, Li, Yang, and Liu]{ReMoDiffuse}
Mingyuan Zhang, Xinying Guo, Liang Pan, Zhongang Cai, Fangzhou Hong, Huirong Li, Lei Yang, and Ziwei Liu.
\newblock {ReMoDiffuse: Retrieval-Augmented Motion Diffusion Model}.
\newblock In \emph{{Proceedings of the IEEE/CVF International Conference on Computer Vision}}, pages 364--373, 2023{\natexlab{b}}.

\bibitem[Zhang et~al.(2025)Zhang, Dabral, Golyanik, Choutas, Alvarado, Beeler, Habermann, and Theobalt]{zhang2025bimart}
Wanyue Zhang, Rishabh Dabral, Vladislav Golyanik, Vasileios Choutas, Eduardo Alvarado, Thabo Beeler, Marc Habermann, and Christian Theobalt.
\newblock Bimart: A unified approach for the synthesis of 3d bimanual interaction with articulated objects.
\newblock \emph{CVPR}, 2025.

\bibitem[Zhao et~al.(2023)Zhao, Liu, Ren, Dai, and Sebe]{zhao2023modiff}
Mengyi Zhao, Mengyuan Liu, Bin Ren, Shuling Dai, and Nicu Sebe.
\newblock Modiff: Action-conditioned 3d motion generation with denoising diffusion probabilistic models, 2023.

\bibitem[Zheng et~al.(2023)Zheng, Zheng, Fang, Liu, and Yi]{zheng2023cams}
Juntian Zheng, Qingyuan Zheng, Lixing Fang, Yun Liu, and Li~Yi.
\newblock Cams: Canonicalized manipulation spaces for category-level functional hand-object manipulation synthesis.
\newblock In \emph{CVPR}, pages 585--594, June 2023.

\bibitem[Zhou et~al.(2019)Zhou, Barnes, Lu, Yang, and Li]{ContinuityOfRotations}
Yi~Zhou, Connelly Barnes, Jingwan Lu, Jimei Yang, and Hao Li.
\newblock On the {{Continuity}} of {{Rotation Representations}} in {{Neural Networks}}.
\newblock In \emph{Proceedings of the {{IEEE}}/{{CVF Conference}} on {{Computer Vision}} and {{Pattern Recognition}}}, pages 5745--5753, 2019.

\end{thebibliography}
\bibliographystyle{plainnat}
}

\end{document}